\documentclass[10pt,twocolumn,letterpaper]{article}

\usepackage{cvpr}              % To produce the CAMERA-READY version
\definecolor{cvprblue}{rgb}{0.21,0.49,0.74}
\usepackage[pagebackref,breaklinks,colorlinks,allcolors=cvprblue]{hyperref}
\newcommand{\holodeck}{\textsc{Holodeck}\xspace}
\newcommand{\holodeckabs}{{\fontfamily{ntxtlf}\selectfont\textsc{Holodeck}}\xspace}
\usepackage{float}
\usepackage{stfloats}

\usepackage{algorithm}
\usepackage{algorithmic}
\usepackage{newfloat}
\usepackage{listings}
\usepackage{caption}

\DeclareCaptionStyle{ruled}{labelfont=normalfont,labelsep=colon,strut=off} % DO NOT CHANGE THIS
\floatstyle{ruled}
\newfloat{listing}{tb}{lst}{}
\floatname{listing}{Listing}

\def\paperID{14372} % *** Enter the Paper ID here
\def\confName{CVPR}
\def\confYear{2026}

\title{\holodeck 2.0: Vision‑Language‑Guided 3D World Generation with Editing}

\author{
Zixuan Bian$^{*}$\\
University of Pennsylvania\\
{\tt\small bianzx@seas.upenn.edu}
\and
Ruohan Ren$^{*}$\\
Cornell University\\
{\tt\small rur4004@med.cornell.edu}
\and
Yue Yang\\
University of Pennsylvania\\
{\tt\small yueyang1@seas.upenn.edu}
\and
Chris Callison-Burch$^{\dagger}$\\
University of Pennsylvania\\
{\tt\small ccb@upenn.edu}
}

\begin{document}
\twocolumn[{%
\maketitle
\vspace{-3em}
\begin{figure}[H]
\hsize=\textwidth
\centering
\includegraphics[width=0.9\textwidth]{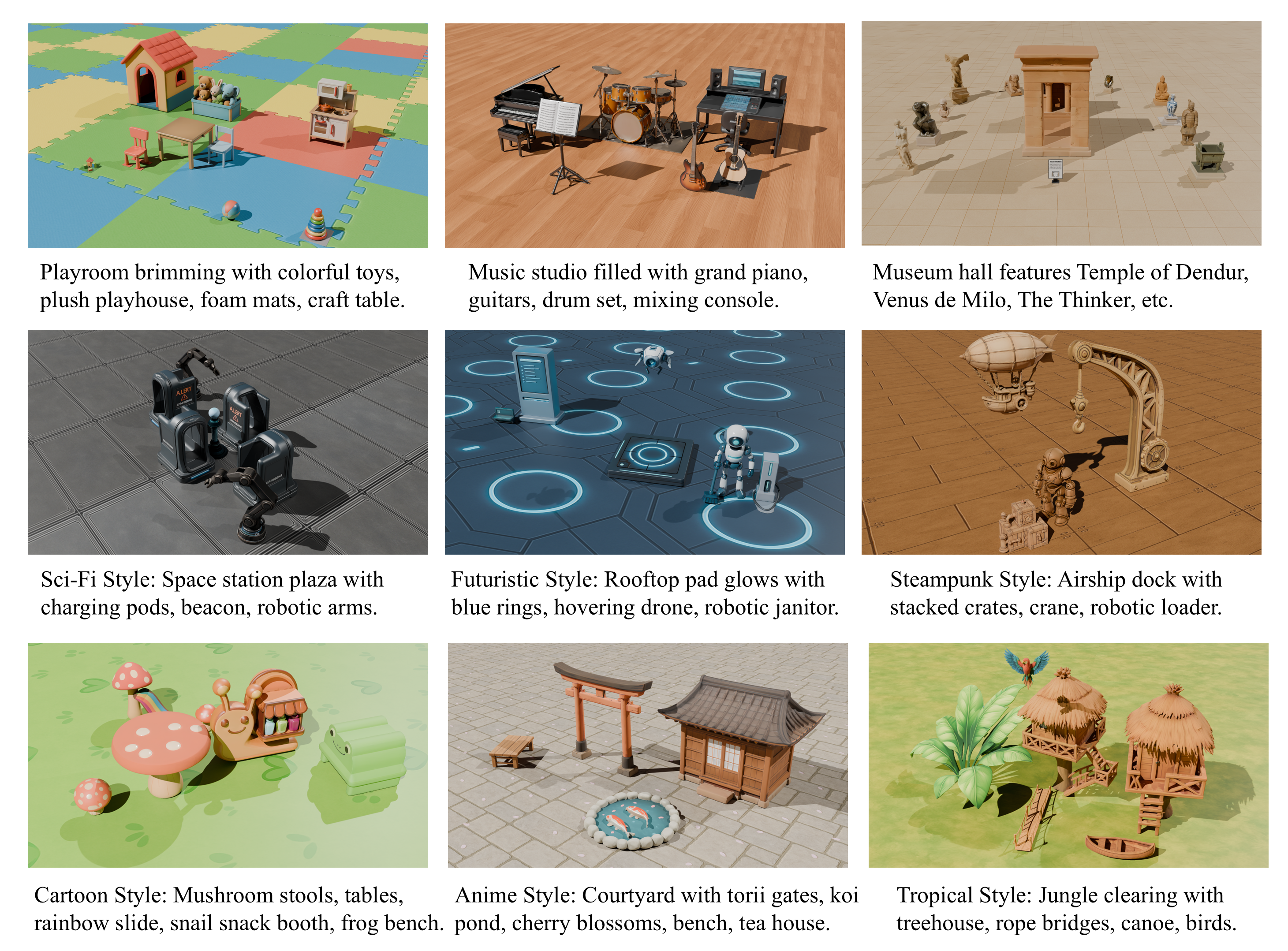}
\caption{\textbf{Examples of stylistically varied 3D scenes generated by \holodeck 2.0.} Captions are shortened versions of the original long inputs (50–300 words). \holodeck 2.0 can faithfully capture fine-grained details from textual descriptions.}
\label{fig1}
\end{figure}
}]
\renewcommand{\thefootnote}{\fnsymbol{footnote}}
\setcounter{footnote}{0}
\footnotetext{$^{*}$Equal contribution.\quad $^{\dagger}$Corresponding author.}
\renewcommand{\thefootnote}{\arabic{footnote}}
\setcounter{footnote}{0}

\begin{abstract}
3D scene generation plays a crucial role in gaming, artistic creation, virtual reality, and many other domains. 
However, current 3D scene design still relies heavily on extensive manual effort from creators, and existing automated methods struggle to generate open-domain scenes or support flexible editing. 
To address those challenges, we introduce \holodeckabs 2.0, an advanced vision-language-guided framework for 3D world generation with support for interactive scene editing based on human feedback. 
\holodeckabs 2.0 can generate diverse and stylistically rich 3D scenes (e.g., realistic, cartoon, anime, and cyberpunk styles) that exhibit high semantic fidelity to fine-grained input descriptions, suitable for both indoor and open-domain environments. 
\holodeckabs 2.0 leverages vision-language models (VLMs) to identify and parse the objects required in a scene and generates corresponding high-quality assets via state-of-the-art 3D generative models. 
Then, \holodeckabs 2.0 iteratively applies spatial constraints derived from the VLMs to achieve semantically coherent and physically plausible layouts. 
Both human and model evaluations demonstrate that \holodeckabs 2.0 effectively generates high-quality scenes closely aligned with detailed textual descriptions, consistently outperforming baselines across indoor and open-domain scenarios. 
Additionally, \holodeckabs 2.0 provides editing capabilities that flexibly adapt to human feedback, supporting layout refinement and style-consistent object edits. 
Finally, we present a practical application of \holodeckabs 2.0 in procedural game modeling to generate visually rich and immersive environments that can boost efficiency in game design. Code is available at \url{https://github.com/bzx20/Holodeck2.0}.
\end{abstract}    
\section{Introduction}
\label{sec:intro}

Generating complex 3D environments from natural language descriptions is a fundamental challenge at the intersection of computer vision, natural language processing, and graphics. 
As virtual worlds become increasingly central to gaming, film production, architectural visualization, and training embodied AI agents, the demand for efficient, high-quality 3D content creation has never been greater.
However, the current practice of 3D scene design still relies predominantly on extensive manual effort by creators. 
Therefore, the task of generating 3D worlds directly from text or images has attracted widespread attention.

Some current approaches focus on directly generating 3D scenes represented by Neural Radiance Fields (NeRFs) or Gaussian splats from text or images \cite{poole2022dreamfusion,lin2023magic3d,chung2023luciddreamer}. 
However, such methods generate scenes that lack separable and manipulable objects, making them difficult to edit directly and limiting their applicability in game design, embodied AI training, etc. 
Another class of methods focuses on selecting suitable 3D objects from asset datasets and then assembling them into a 3D scene \cite{paschalidou2021atiss,yang2024Holodeck,hu2024scenecraft,aguina2024openuniverse}. 
For example, \holodeck \cite{yang2024Holodeck} uses Large Language Models (LLMs) to parse from the input text the objects required for a scene, then retrieves 3D assets from datasets such as Objaverse \cite{deitke2023objaverse} based on visual similarities using CLIP \cite{radford2021clip} scores, and finally arranges them into a layout.

Specifically, existing approaches face three critical limitations: (1) asset-based methods are constrained by the \textbf{quality and diversity of available 3D assets}; even with large datasets like Objaverse-XL \cite{deitke2023objaverse-xl}, they still suffer from inconsistent quality and missing textures; 
(2) current systems struggle with \textbf{fine-grained style control}, unable to maintain consistent artistic styles (e.g., anime, cyberpunk) across all objects in a scene; 
and (3) most methods focus exclusively on \textbf{indoor environments}, lacking the flexibility to generate open-world scenarios like fantasy landscapes or futuristic cities.

Recent progress in vision-language models (VLMs) \cite{llava,molmo} and 3D generative models \cite{poole2022dreamfusion} presents an unprecedented opportunity to address these limitations. VLMs now possess sophisticated spatial reasoning capabilities \cite{chen2024spatialvlm} that can parse complex scene descriptions and generate physically plausible arrangements of objects. 
Simultaneously, state-of-the-art 3D generative models like Hunyuan3D \cite{hunyuan3d2025hunyuan3d2.1} can create high-quality, stylistically consistent assets from text or image prompts. 
However, these advances have not yet been effectively integrated into a unified 3D scene generation framework.

Facilitated by 3D asset generative models and the inference capabilities of VLMs, \holodeck 2.0 can produce high-quality 3D scenes with internal stylistic coherence for both indoor settings (e.g., music studios, playrooms) and open domains (e.g., airship docks, courtyards), as shown in Figure \ref{fig1}. 
Meanwhile, \holodeck 2.0 can generate the same type of 3D scenes in different styles, such as realistic, cartoon, and steampunk. 
The objects in \holodeck 2.0 scenes exhibit high quality and closely match their descriptions, even when those descriptions are stylized, imaginative, and richly detailed, for example, ``A cartoon-style snail-shaped snack booth offers glowing drinks'' as shown in Figure \ref{fig1}. 
In terms of spatial reasoning, \holodeck 2.0 fully leverages the spatial understanding capabilities of VLMs, iteratively generating and applying spatial constraints to produce layouts that match both textual and visual descriptions.

We demonstrate the effectiveness of \holodeck 2.0 through a comprehensive evaluation that encompasses large-scale human assessments, CLIP-based perceptual score, physical plausibility measures, and spatial-semantic alignment scores.
Across 70 test cases spanning both indoor and open-domain scenes, \holodeck 2.0 consistently outperforms the baselines, with its advantage being particularly pronounced in open-domain scenes.

In addition, we demonstrate \holodeck 2.0's practical applicability through direct integration with commercial game engines. 
We showcase how a complex museum scene, complete with artifacts such as the Temple of Dendur, Venus de Milo, and The Thinker, can be generated from detailed textual descriptions and seamlessly imported into Unreal Engine with full interactivity. 
This shows how \holodeck 2.0 can accelerate 3D content creation workflows for game development and other applications.

In summary, our paper contributes as follows:
\begin{enumerate}
\item We present a unified framework that uniquely combines VLM-guided scene decomposition with generative 3D asset creation, utilizing VLMs across visual reference generation, object parsing, spatial constraint generation, and iterative layout refinement.
\item We demonstrate state-of-the-art performance on both indoor and open-domain scene generation with consistent artistic styles (realistic, cartoon, anime, cyberpunk, etc.) through extensive human and automated evaluations.
\item We introduce an interactive editing system that leverages generative models for style-consistent object replacement and addition, beyond traditional layout adjustments.
\item We validate downstream applications through integrating with commercial game engines and demonstrate significant quality improvements over baselines across perceptual, physical, and spatial-semantic metrics.
\end{enumerate}

\begin{figure*}[htbp]
\centering
\includegraphics[width=1.0\textwidth]{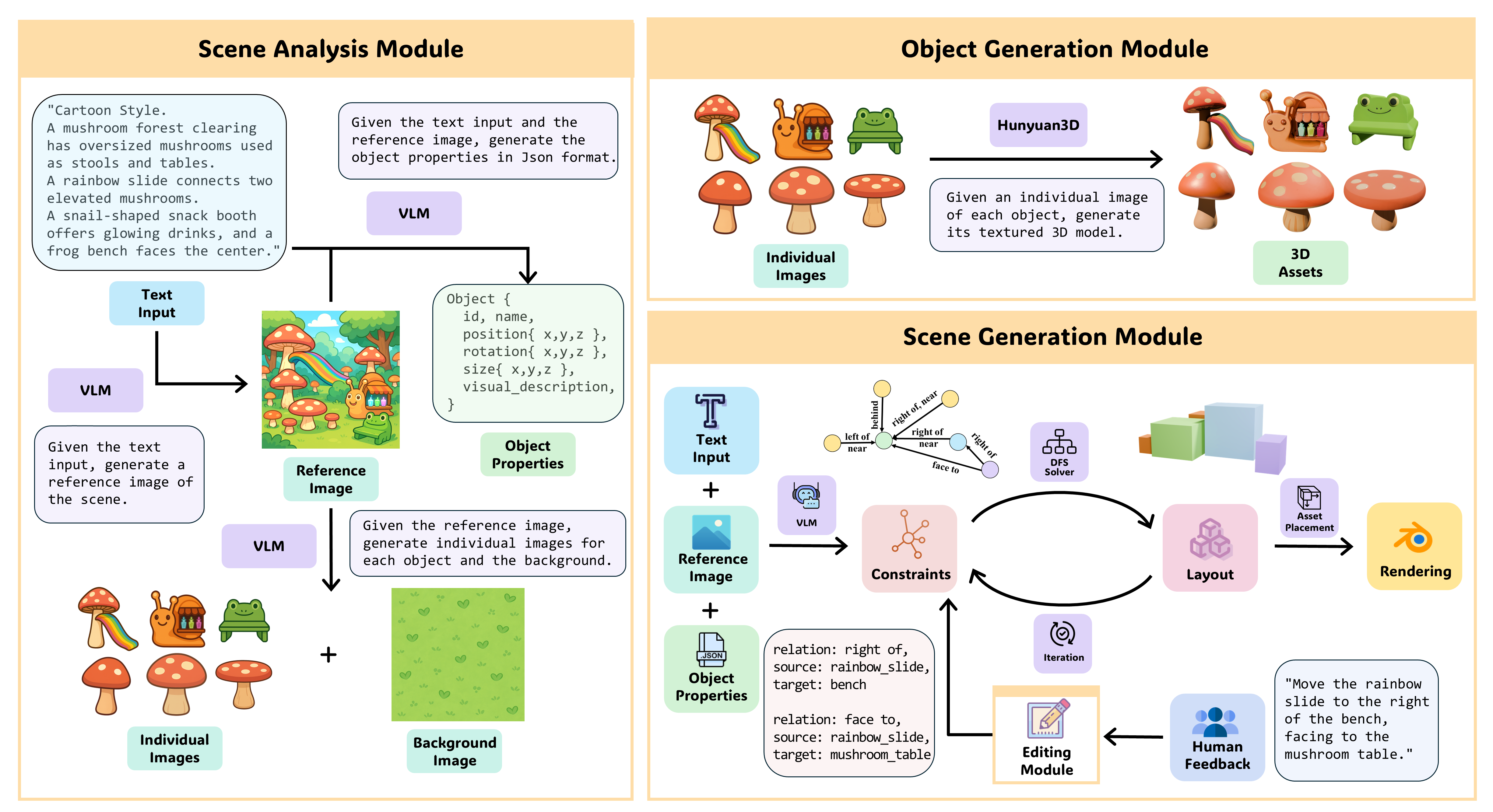} 
% \caption{\textbf{Overview of \holodeck 2.0}. Given a text input, \holodeck 2.0 generates 3D scenes via three modules, with an optional editing module for interactive updates based on human language feedback. \yy{briefly explain the three modules (input/output) here}}
\caption{\textbf{Overview of \holodeck 2.0}. Given a text input, \holodeck 2.0 generates 3D scenes via three modules: (1) \textit{Scene Analysis Module} takes text as input and outputs object properties in JSON format, along with a reference image, individual object images, and a background image; (2) \textit{Object Generation Module} takes individual object images as input and outputs textured 3D assets; (3) \textit{Scene Generation Module} takes text, reference image, and object properties as input, and outputs the final 3D scene, with an optional editing module for human feedback.}
\label{fig2}
\end{figure*}

\section{Related Work} \label{sec:related}
\textbf{Language-Guided 3D Scene Generation.}
Early work in this area is primarily symbolic and rule‑based, requiring extensive engineering effort and exhibiting very limited generalization ability, for example, Wordseye \cite{coyne2001wordseye} and other systems \cite{chang2014learningspatial,ma2018language3d}. With the emergence of LLMs and VLMs, open‑vocabulary 3D scene generation has become possible, greatly enhancing the flexibility of scene creation \cite{yang2024Holodeck, hu2024scenecraft, zhang2025scenelanguage, lin2025partcrafter, ccelen2024I-design, fu2024anyhome, ocal2024sceneteller, deng2025global-local-tree-vlm, sun2025hierarchically, sun20253d-gpt}. Among them, \holodeck \cite{yang2024Holodeck} employs LLMs to get the object descriptions that should appear in a scene based on the input text, and to generate spatial constraints among those objects. SceneCraft \cite{hu2024scenecraft} also leverages LLMs and VLMs in a similar way, but it generates Blender code and iteratively optimizes it to produce 3D scenes.
However, both of them rely on CLIP \cite{radford2021clip} to retrieve 3D objects from asset libraries such as Objaverse \cite{deitke2023objaverse} and TurboSquid \cite{c:TurboSquid}, and the variety and quality of these assets somewhat limit their capability to generate open‑world 3D scenes. Currently, larger or better‑curated 3D asset libraries have emerged \cite{deitke2023objaverse-xl, lin2025objaverse++}. However, they still suffer from overall low asset quality, missing textures, and limited variety and styles of objects. For 3D scene spatial layout, LayoutGPT \cite{feng2023layoutgpt} is a method that directly generates 3D object layout relationships via LLM prompting, but often neglects physical plausibility. LayoutVLM \cite{sun2025layoutvlm} is another method that uses VLMs to generate 3D layouts by formulating objective functions that can be differentiably optimized. For 3D scene editing, recent works \cite{c:BlenderGPT, huang2024blenderalchemy, gu2025blendergym} have applied LLMs or VLMs to editing, though they still face challenges in accurately managing object placement and layout.
\medbreak
\noindent
\textbf{Text-to-3D Diffusion Models.}
The emergence of diffusion models \cite{sohl2015deepthermo, ho2020ddpm, peebles2023dit, rombach2022stablediffu} marks a significant advancement in the field of generative modeling. Current methods leverage diffusion models to directly generate 3D objects or scenes represented by Neural Radiance Fields (NeRFs) or Gaussian splats from text or images \cite{poole2022dreamfusion, lin2023magic3d, li2024dreamscene, po2024compositional}. However, such methods generate scenes that lack discrete, manipulable objects, making them difficult to edit directly and unsuitable for applications like game design or embodied AI training. When focusing solely on generating individual 3D assets, diffusion models offer significant advantages. Recent methods have achieved breakthroughs in generating high‑quality 3D assets \cite{xiang2025TRELLIS, zhang20233dshape2vecset, zhang2024clay, zhao2023michelangelo, li2024craftsman3d, ren2024xcube, wu2024direct3d}, with Hunyuan3D \cite{hunyuan3d2025hunyuan3d2.1, lai2025hunyuan3d2.5} being a typical example.

\section{\holodeck 2.0}
\holodeck 2.0 is a vision-language-guided system, facilitated by VLMs and 3D generative models, that can produce and edit diverse, customizable, and stylistically varied 3D scenes.

As shown in Figure \ref{fig2}, \holodeck 2.0 systematically generates 3D scenes from text utilizing the following specialized modules: (1) \textbf{Scene Analysis Module} parses the object properties and background information that should appear in the scene from input text; (2) \textbf{Object Generation Module} leverages 3D asset generative models to produce 3D assets from quality‑controlled object images; (3) \textbf{Scene Generation Module} iteratively generates semantically and physically coherent object layouts based on spatial constraints; (4) \textbf{Scene Editing Module} automatically applies personalized adjustments to the scene based on human feedback.

In the following, we describe how each module of \holodeck 2.0 operates, specifically illustrating how it coordinates VLMs and 3D asset generative models to accomplish scene generation and editing. Detailed information about \holodeck 2.0 is also available in the supplement.

\begin{figure*}[htbp]
\centering
\includegraphics[width=0.95\textwidth]{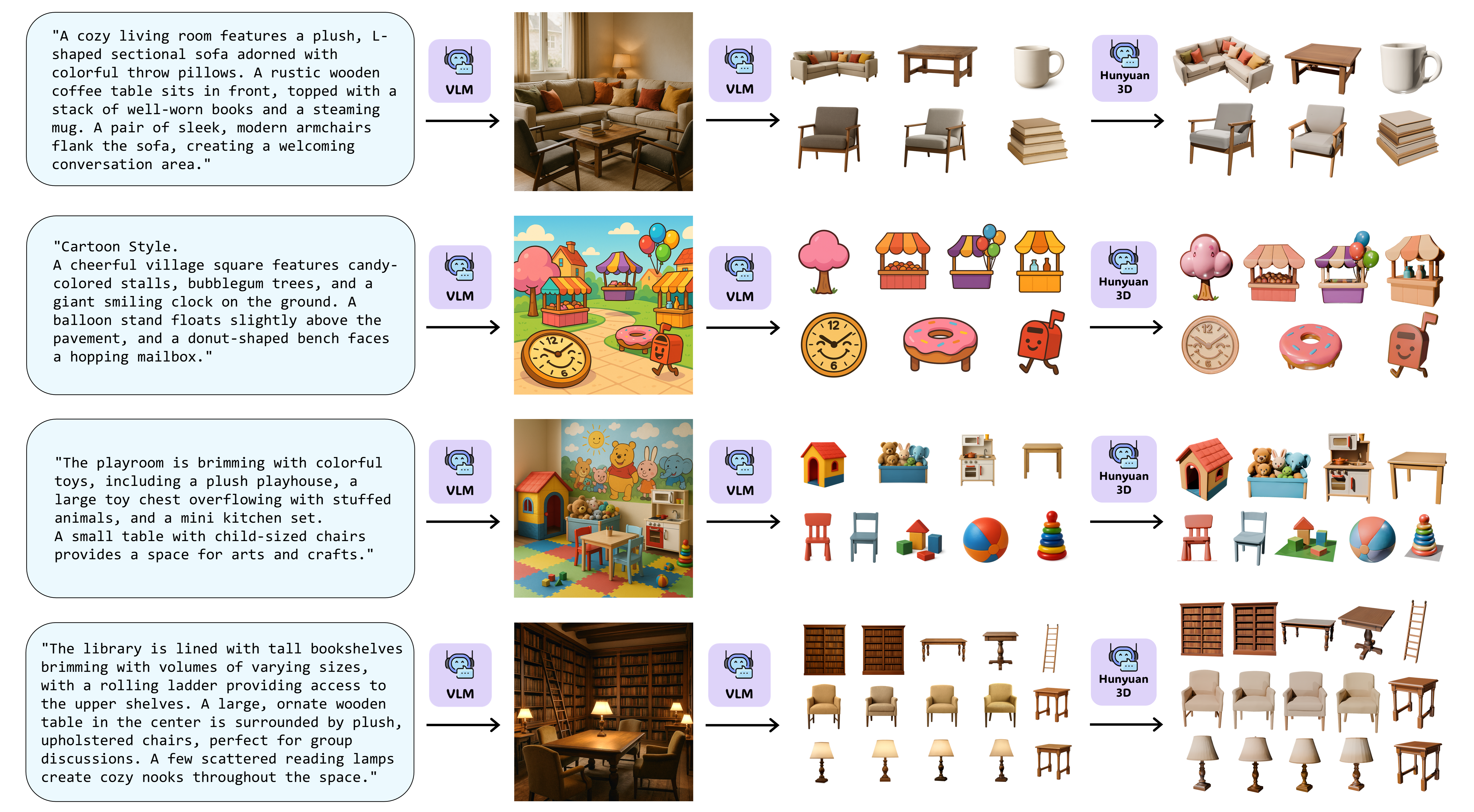}
\caption{Examples of the \textbf{Scene Analysis Module} and the \textbf{Object Generation Module}. \holodeck 2.0 can generate customized, stylistically diverse 3D objects that precisely match fine-grained textual descriptions.}
\label{fig3}
\end{figure*}

\subsection{Scene Analysis}
\noindent The \textbf{Scene Analysis Module}, shown in the first panel of Figure \ref{fig2}, is designed to parse object properties that should appear in the scene from input text. 
\smallbreak
\noindent \textit{Text to Scene Image.} Given a text input, the Scene Analysis Module uses GPT-Image-1 to generate a corresponding reference image. This image generation process is vital for the subsequent creation of uniform‑style assets. Moreover, visual inputs can substantially enhance the spatial awareness of large reasoning models. In Figure \ref{fig3}, we present several examples of scene reference images generated from text.
\smallbreak
\noindent \textit{Scene Image to Object Images.} After obtaining the scene image, the Scene Analysis Module employs GPT-o3 \cite{GPTo3} to infer object properties (name, position, rotation, size, visual description, etc.) from the text input and its corresponding scene reference image. Then we prompt GPT-Image-1 to generate a clear, transparent frontal image of each object appearing in the scene image. This avoids the issues common to current image segmentation algorithms \cite{kirillov2023segmentanything, ren2024groundedSAM}, such as inaccurate segmentation, blurry edges, and difficulty handling occluded objects. After generating individual object images, we use GPT-o3 to perform quality control, removing smaller-object images redundantly embedded in larger-object images (e.g., removing the separate faucet when it is already present in the bathtub image). In practice, this approach generates accurate, high‑quality object images from the reference image as shown in Figure \ref{fig3}. Subsequently, we generate the scene's background image based on the reference image, focusing primarily on the ground texture.

\subsection{Object Generation}
\noindent The \textbf{Object Generation Module}, shown in the second panel of Figure \ref{fig2}, utilizes 3D asset generative models to generate 3D assets from quality-controlled object images. For 3D asset generation, image-to-3D is faster and offers better style control than direct text-to-3D. After obtaining the object images from the Scene Analysis Module, we use a locally deployed Hunyuan3D 2.1  \cite{hunyuan3d2025hunyuan3d2.1} to generate the corresponding 3D asset for each object. In practice, compared to searching for 3D objects in Objaverse, generative models can produce higher‑quality 3D assets that more faithfully match the descriptions (Figure \ref{fig3}), for example, ``a plush, L-shaped sectional sofa adorned with colorful throw pillows" and ``a cartoon-style donut-shaped bench".

\subsection{Scene Generation}
\noindent The \textbf{Scene Generation Module}, shown in the third panel of Figure \ref{fig2}, can generate semantically and physically coherent object layouts through iterative cycles between generated spatial constraints and a Depth‑First-Search (DFS) solver.

\noindent \textit{Spatial Constraint Generation}. The Scene Generation Module first employs GPT‑o3 to produce a series of spatial constraints between objects based on the text input, the scene reference image, and object properties. We define a total of ten spatial relations, which are divided into the following four categories: (1) Relative: \textit{in front of}, \textit{behind}, \textit{left of}, \textit{right of}, \textit{side of}; (2) Distance: \textit{near}, \textit{far}; (3) Vertical: \textit{on}, \textit{above} and (4) Rotation: \textit{face to}. Their mathematical definitions are provided in the supplement. As shown in Figure \ref{fig4}, reasoning models (e.g., GPT-o3) can generate realistic and spatially plausible constraints for objects.
\smallbreak
\noindent \textit{Layout Generation}. For layout generation, we employ a DFS solver to compute object placements based on object properties and spatial constraints between objects. 
We define mathematical formulations to enforce each spatial constraint (for example, ``face to" means that the angle between the source object’s orientation and the line connecting the centers of the two objects is less than 10 degrees). 
Objects should satisfy these spatial constraints while avoiding physical collisions. 
To fully leverage GPT-o3's spatial reasoning capabilities, we prioritize the initial object positions inferred by the model. 
It is worth noting that when the DFS solver encounters spatial constraints that are difficult to satisfy simultaneously, the Scene Generation Module re-prompts GPT-o3 using the DFS solver's feedback to update the objects' spatial constraints. 
The DFS solver and the reasoning model then form an iterative loop until a valid layout is found. 
This approach significantly enhances \holodeck 2.0’s layout capabilities. Figure \ref{fig4} presents examples of position plots for the bounding boxes produced by the solver.

\begin{figure}[t]
\centering
\includegraphics[width=0.5\textwidth]{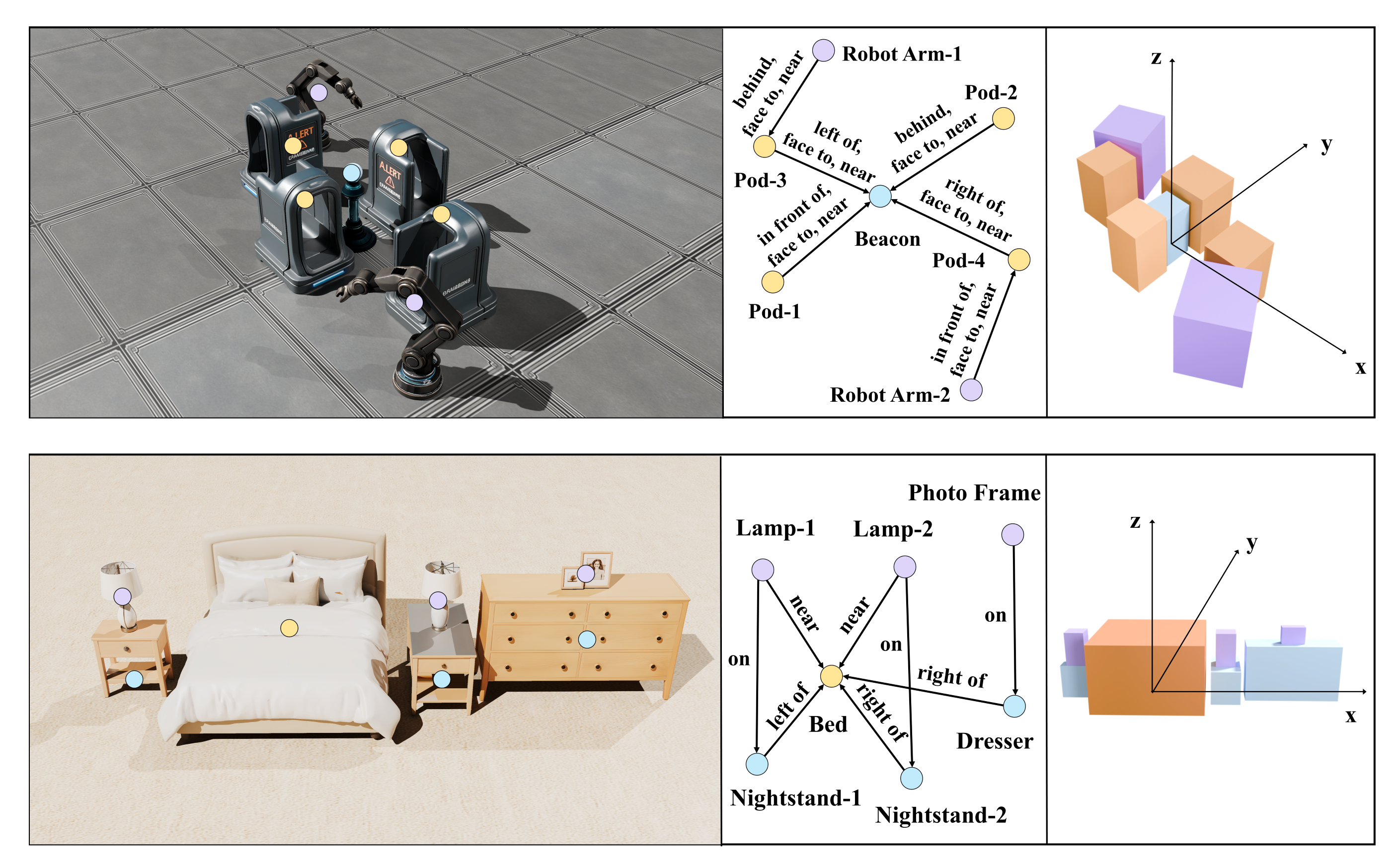} % Reduce the figure size so that it is slightly narrower than the column. Don't use precise values for figure width.This setup will avoid overfull boxes.
\caption{Example outputs of the \textbf{Scene Generation Module}. \holodeck 2.0 can generate appropriate spatial constraints and optimal positioning for object layouts.}
\label{fig4}
\end{figure}

\subsection{Scene Editing}
\begin{figure}[t]
\centering
\includegraphics[width=0.5\textwidth]{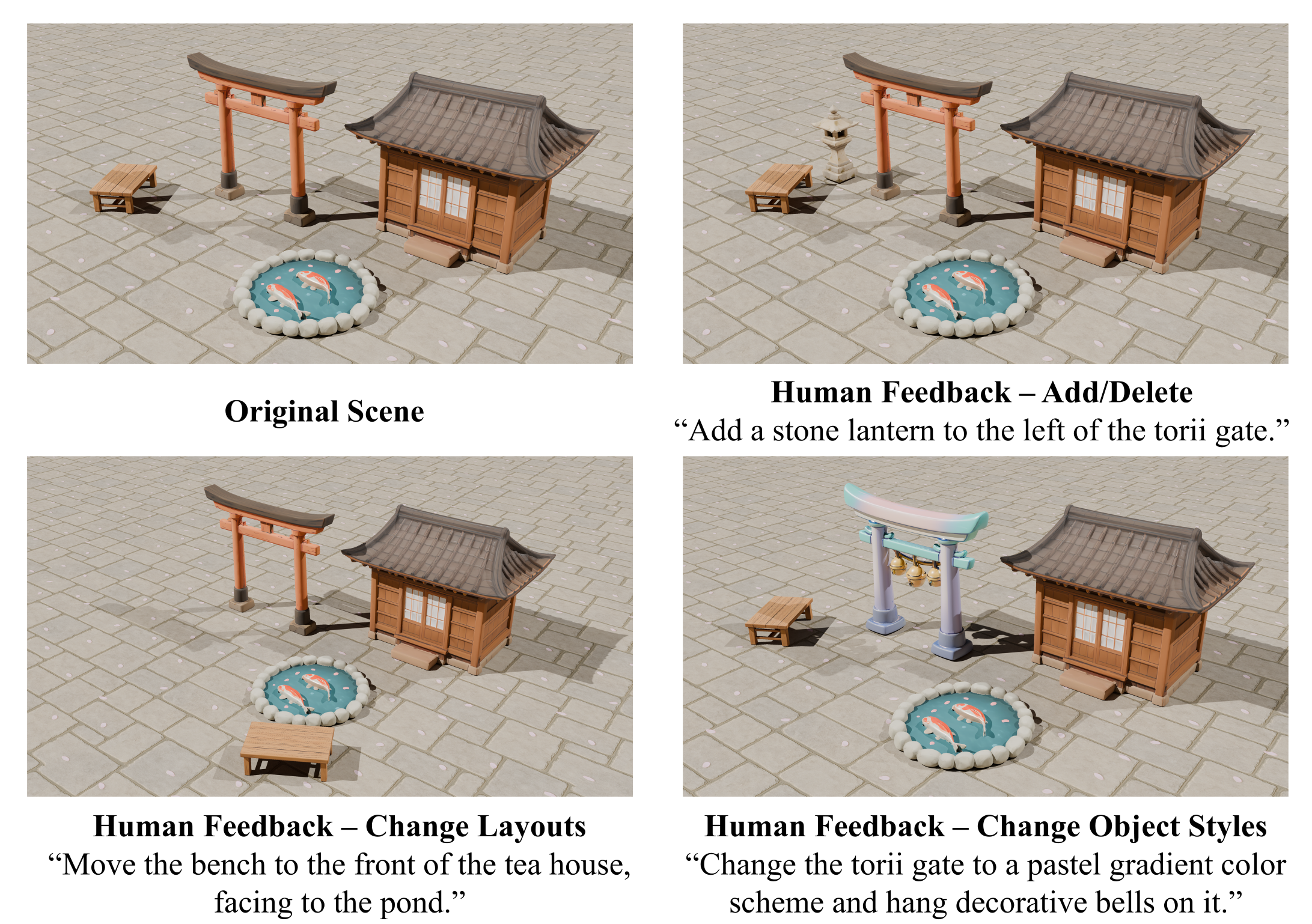} % Reduce the figure size so that it is slightly narrower than the column. Don't use precise values for figure width.This setup will avoid overfull boxes.
\caption{The \textbf{Scene Editing Module} of \holodeck 2.0 can add/delete objects, change layouts, and change object styles based on human language feedback.}
\label{fig5}
\end{figure}

\noindent The \textbf{Scene Editing Module}, shown as a submodule in the third panel of Figure \ref{fig2}, is designed to automatically apply personalized adjustments to the scene based on user language feedback when needed. It can operate at both the scene layout level and the 3D asset level (Figure \ref{fig5}).

\noindent \textit{Scene Layout Level}. If users want to personalize the generated scene layout after it has been created, they can provide a natural-language edit instruction. Then, it is incorporated into the original workflow to regenerate the spatial constraints between objects, and the DFS solver iteratively computes the updated layout.

\noindent \textit{3D Asset level}. Utilizing 3D generative models, we support asset‑level editing. Because our scenes are built by assembling individual 3D assets, we can easily delete objects in response to user language feedback. Users can also replace objects, for example, ``Change the torii gate to a pastel gradient color scheme and hang decorative bells on it." The Scene Editing Module then uses that language feedback to regenerate a style‑consistent 3D asset for the specified object and substitute it into the scene. When adding new objects, the module regenerates suitable 3D assets and recalculates spatial constraints to maintain coherent scene layouts.

\section{Experiments}
% To systematically evaluate the quality of 3D scenes generated by \holodeck 2.0, we employ a combination of human evaluations, CLIP‑based scoring and VLM-based scoring. \holodeck \cite{yang2024Holodeck} is a representative method for 3D scene generation that retrieves objects from 3D asset libraries and is one of the most commonly used baselines in recent work. Although it primarily focuses on indoor scene generation, its open‑vocabulary text input makes it applicable to open‑domain scenes to some extent. Therefore, we choose to directly compare with the \holodeck method. Specifically, we generated 70 test cases spanning both indoor and open‑domain scenes. Our comparisons show that \holodeck 2.0 outperforms the baselines in both indoor and open‑domain settings, with the advantage being more pronounced in open‑domain scenarios. In addition, we perform ablation studies on the Scene Generation Module. We further compare the Scene Editing Module with the baseline method, BlenderAlchemy \cite{huang2024blenderalchemy} and BlenderGym \cite{gu2025blendergym}. The detailed results are shown below.
To systematically evaluate the quality of 3D scenes generated by \holodeck 2.0, we employ a combination of human evaluations, CLIP‑based scoring and VLM-based scoring in Section \ref{sec:comparative}. Specifically, we generate 70 test cases spanning both indoor and open‑domain scenes. Our comparisons show that \holodeck 2.0 outperforms the baselines in both indoor and open‑domain settings, with the advantage being more pronounced in open‑domain scenarios. In Section \ref{sec:ablate}, we perform ablation studies on the Scene Generation Module. In Section \ref{sec:edit}, we further compare the Scene Editing Module with the baseline methods, BlenderAlchemy \cite{huang2024blenderalchemy} and BlenderGym \cite{gu2025blendergym}. In Section \ref{sec:app}, we additionally demonstrate \holodeck 2.0’s practical applicability through direct integration with commercial game engines. The detailed results are shown below.

\begin{table*}[htbp]
  \caption{\textbf{Benchmarking performance on indoor and open-domain scenes.}
  We report perceptual, physical, and spatial-semantic metrics. GraphDreamer's metrics are limited due to its implicit representation (see supplement). }
  \label{table1}
  \centering
  \small
  \setlength{\tabcolsep}{1pt}
  \begin{tabular*}{\textwidth}{@{\extracolsep{\fill}}l ccc cc ccc ccc cc ccc@{}}
    \toprule
    & \multicolumn{8}{c}{\textbf{Indoor}} & \multicolumn{8}{c}{\textbf{Open-domain}} \\
    \cmidrule(lr){2-9} \cmidrule(lr){10-17}
    \textbf{Method}
    & \multicolumn{3}{c}{\textbf{Perceptual}} & \multicolumn{2}{c}{\textbf{Physical}} & \multicolumn{3}{c}{\textbf{Spatial-Semantic}}
    & \multicolumn{3}{c}{\textbf{Perceptual}} & \multicolumn{2}{c}{\textbf{Physical}} & \multicolumn{3}{c}{\textbf{Spatial-Semantic}} \\
    \cmidrule(lr){2-4} \cmidrule(lr){5-6} \cmidrule(lr){7-9}
    \cmidrule(lr){10-12} \cmidrule(lr){13-14} \cmidrule(lr){15-17}
      & CLIP$\uparrow$ & HA$\uparrow$ & HS$\uparrow$ & CF(\%)$\uparrow$ & CN$\downarrow$ & Pos.$\uparrow$ & Rot.$\uparrow$ & PSA$\uparrow$
      & CLIP$\uparrow$ & HA$\uparrow$ & HS$\uparrow$ & CF(\%)$\uparrow$ & CN$\downarrow$ & Pos.$\uparrow$ & Rot.$\uparrow$ & PSA$\uparrow$ \\
    \midrule
    \holodeck        & 0.256 & 4.70 & 3.99 & 51.43 & 1.89 & 67.83 & 69.06 & 38.04
                    & 0.257 & 5.55 & 4.80 & 37.14 & 3.51 & 72.29 & 68.86 & 27.67 \\
    PartCrafter     & 0.259 & 5.02 & 4.55 & 71.42 & 0.29 & 80.03 & 78.91 & 58.76
                    & 0.231 & 3.09 & 2.43 & \textbf{100.00} & \textbf{0.00} & 57.06 & 61.11 & 64.00 \\
    Scene Language   & 0.210 & 2.81 & 2.76  & 47.37 & 1.53 & 74.09 & 65.73 & 37.06
                    & 0.216 & 2.19 & 2.10 & 36.00 & 2.56 & 65.33 & 71.15 & 27.71 \\
    GraphDreamer    & 0.170 & -- & -- & --   & --   & 15.50 & 12.25 & --
                    & 0.226 & -- & -- & --   & --   & 26.17 & 26.67 & -- \\
    \midrule
    \textbf{Ours}   & \textbf{0.297} & \textbf{8.31} & \textbf{7.86} & \textbf{97.14} & \textbf{0.11} & \textbf{81.03} & \textbf{82.86} & \textbf{82.67}
                    & \textbf{0.307} & \textbf{8.07} & \textbf{7.84} & 97.14 & 0.86 & \textbf{84.43} & \textbf{83.63} & \textbf{86.00} \\
    \bottomrule
  \end{tabular*}
\end{table*}

\begin{table*}[t]
  \caption{\textbf{Ablation study results.} \textit{HP} denotes human preference when comparing two scenes with different layouts: w/o Layout Solver and our full method. (Tied cases are split equally between both methods.) }
  \label{table2}
  \centering
  % \normalsize
  \small
  \setlength{\tabcolsep}{1pt}
  \begin{tabular*}{\textwidth}{@{\extracolsep{\fill}}l cc cc ccc cc cc ccc@{}}
    \toprule
    & \multicolumn{7}{c}{\textbf{Indoor}} & \multicolumn{7}{c}{\textbf{Open-domain}} \\
    \cmidrule(lr){2-8} \cmidrule(lr){9-15}
    \textbf{Method}
    & \multicolumn{2}{c}{\textbf{Perceptual}} & \multicolumn{2}{c}{\textbf{Physical}} & \multicolumn{3}{c}{\textbf{Spatial-Semantic}}
    & \multicolumn{2}{c}{\textbf{Perceptual}} & \multicolumn{2}{c}{\textbf{Physical}} & \multicolumn{3}{c}{\textbf{Spatial-Semantic}} \\
    \cmidrule(lr){2-3} \cmidrule(lr){4-5} \cmidrule(lr){6-8}
    \cmidrule(lr){9-10} \cmidrule(lr){11-12} \cmidrule(lr){13-15}
      & CLIP$\uparrow$ & HP(\%)$\uparrow$ & CF(\%)$\uparrow$ & CN$\downarrow$ & Pos.$\uparrow$ & Rot.$\uparrow$ & PSA$\uparrow$
      & CLIP$\uparrow$ & HP(\%)$\uparrow$ & CF(\%)$\uparrow$ & CN$\downarrow$ & Pos.$\uparrow$ & Rot.$\uparrow$ & PSA$\uparrow$ \\
    \midrule
    \textit{w/o} Layout Solver     & 0.289 & 24.63 & 54.29 & 1.23 & 77.83 & 76.11 & 
               45.30 & 0.306 & 23.70 & 42.86 & 1.89 & 79.09 & 77.83 & 36.53 \\
    \midrule
    \textbf{Ours}   & \textbf{0.297} & \textbf{75.37} & \textbf{97.14} & \textbf{0.11} & \textbf{81.03} & \textbf{82.86} & \textbf{82.67}
                    & \textbf{0.307} & \textbf{76.30} & \textbf{97.14} & \textbf{0.86} & \textbf{84.43} & \textbf{83.63} & \textbf{86.00} \\
    \bottomrule
  \end{tabular*}
\end{table*}

\subsection{Comparative Analysis }
\label{sec:comparative}
\textbf{Metrics.} We evaluate generated 3D scenes along three complementary dimensions: \textit{perceptual}, \textit{physical}, and \textit{spatial-semantic}. A detailed description of all metrics is provided below:

% \noindent\textit{Perceptual Quality}
% \begin{itemize}[noitemsep, left=1em]
%     \item CLIP similarity: Measures similarity between rendered scene and input text (using CLIP ViT-B/32 model).
%     \item Human rating at the Asset level (HA): Assesses realism and fidelity of individual 3D assets.
%     \item Human rating at the Scene level (HS): Evaluates holistic scene quality and consistency with text.
% \end{itemize}

% \noindent\textit{Physical Plausibility}
% \begin{itemize}[noitemsep, left=1em]
%     \item Collision-Free ratio (CF): Fraction of scenes without collisions between objects.
%     \item Collision Number (CN): Average number of collision pairs per scene.
% \end{itemize}

% \noindent\textit{Spatial-Semantic Consistency}
% \begin{itemize}[noitemsep, left=1em]
%     \item Positional Coherency (Pos.): Measures how well object positions match text-indicated spatial relations.
%     \item Rotational Coherency (Rot.): Measures consistency of object orientations relative to the textual descriptions.
%     \item Physically-grounded Semantic Alignment (PSA): Holistic evaluation combining spatial and physical grounding \cite{sun2025layoutvlm}.
% \end{itemize}

\noindent \textbf{(1) Perceptual Quality}
\begin{itemize}[noitemsep, left=1em]
    \item \textit{CLIP score}: Text–image similarity between scene rendering and input text, computed using CLIP (ViT-B/32).
    \item \textit{Human rating at the Asset level (HA)}: Human evaluation of individual 3D assets' realism and fidelity.
    \item \textit{Human rating at the Scene level (HS)}: Human evaluation of overall scene plausibility and consistency with the input text.
\end{itemize}

\noindent \textbf{(2) Physical Plausibility}
\begin{itemize}[noitemsep, left=1em]
    \item \textit{Collision-Free ratio (CF)}: The percentage of scenes without collisions between objects.
    \item \textit{Collision Number (CN)}: Average number of collision pairs per scene.
\end{itemize}

\noindent \textbf{(3) Spatial-Semantic Consistency}
\begin{itemize}[noitemsep, left=1em]
    \item \textit{Positional Coherency (Pos.)}: Measures how well object positions match the spatial relations indicated by the text.
    \item \textit{Rotational Coherency (Rot.)}: Measures how well object orientations match the directional relations indicated by the text.
    \item \textit{Physically-grounded Semantic Alignment (PSA)}: A holistic metric that evaluates semantic spatial correctness together with physical feasibility.
\end{itemize}

\smallbreak
The complete formulation of those metrics can be found in the supplement.

% \smallbreak
% \noindent \textit{Perceptual quality.} We report the CLIP similarity (using CLIP ViT-B/32) \cite{radford2021clip} between each rendered scene and its input text, along with human evaluation scores at two levels: \textit{Human rating at the Asset level (HA)}, assessing the realism and fidelity of individual 3D assets; and \textit{Human rating at the Scene level (HS)}, evaluating the holistic quality and consistency of the generated scene with respect to the textual description. 
% \smallbreak
% \noindent \textit{Physical plausibility.} We use \textit{Collision-Free ratio (CF)} and \textit{average Collision Number (CN)}, which measure how often and how severely objects collide within each scene. 
% \smallbreak
% \noindent \textit{Semantic Consistency.}
% We further adopt three spatial-semantic metrics: \textit{Positional Coherency (Pos.)}, \textit{Rotational Coherency (Rot.)}, and the \textit{Physically-grounded Semantic Alignment (PSA)}, as defined in \cite{sun2025layoutvlm}, to evaluate how well the generated objects' positions, orientations, and overall scene layouts align with textual spatial relations.

% The complete formulation of those metrics can be found in the supplement.

\medbreak
\noindent\textbf{Baselines.}
We compare \holodeck 2.0 against four representative 3D scene generation approaches: \holodeck \cite{yang2024Holodeck}, PartCrafter \cite{lin2025partcrafter}, Scene Language \cite{zhang2025scenelanguage}, and GraphDreamer \cite{gao2024graphdreamer}. 
\holodeck \cite{yang2024Holodeck} is a representative method for 3D scene generation that retrieves objects from 3D asset libraries. 
PartCrafter \cite{lin2025partcrafter} jointly generates multiple 3D parts and objects from a single image using compositional latent diffusion transformers. 
Scene Language \cite{zhang2025scenelanguage} represents scenes with programs, words, and embeddings inferred from language models for 3D/4D generation. 
GraphDreamer \cite{gao2024graphdreamer} takes a scene graph as input and generates a compositional 3D scene where each object is fully disentangled.
\medbreak
\noindent\textbf{Setup.} We construct a benchmark of 70 text inputs, comprising 35 indoor scenes and 35 open-domain scenes. (See the supplement for examples of detailed descriptions, ranging from 50--300 words). Indoor text inputs describe environments such as bedrooms, gyms, classrooms, and laboratories, while open-domain text inputs include plazas, city streets, forests, and theme-park settings. 

To accommodate different input requirements across methods, we apply necessary preprocessing while maintaining fairness. Since PartCrafter \cite{lin2025partcrafter} requires image inputs rather than text, we use GPT-Image-1 to convert text descriptions into reference images. For GraphDreamer \cite{gao2024graphdreamer}, which requires object-level descriptions as input, we use preprocessing scripts to extract and format them from the full scene descriptions.

For each text input, we generate corresponding scenes using all four baseline methods and \holodeck 2.0. Human evaluation is conducted with responses from undergraduate and graduate students, who rate each rendered scene on a 1--10 scale at both the asset and scene levels. Automated metrics (\textit{CLIP}, \textit{CF}, \textit{CN}, \textit{Pos.}, \textit{Rot.}, \textit{PSA}) are computed directly from the rendered outputs using a unified evaluation script across all methods.
\medbreak
\noindent \textbf{Results.} The indoor and open-domain surveys are completed by 183 and 268 annotators, respectively. Table \ref{table1} reports quantitative results across all metrics, including human assessments, the CLIP-based perceptual score, physical plausibility measures, and spatial-semantic alignment scores. 

Overall, \holodeck 2.0 achieves the best performance across nearly all metrics in both indoor and open-domain settings. For perceptual quality,  \holodeck2.0 attains the highest \textit{CLIP score} (0.297 for indoor and 0.307 for open-domain) and substantially higher human evaluation scores at both asset and scene levels. Physical plausibility also improves significantly, with  \holodeck 2.0 maintaining a near-perfect \textit{CF} (97.14\%) while keeping the \textit{CN} low (0.11 for indoor and 0.86 for open-domain). In terms of spatial-semantic coherence, \holodeck 2.0 yields the best positional and rotational consistency and the highest \textit{PSA} scores (82.67 for indoor and 86.00 for open-domain), indicating precise alignment between generated layouts and textual spatial relations.

Among the baselines, PartCrafter performs the best after  \holodeck 2.0, particularly in physical stability, while \holodeck and Scene~Language show moderate perceptual alignment but suffer from frequent object collisions. 
GraphDreamer shows lower performance than others.
Overall, these results demonstrate that  \holodeck 2.0 achieves superior perceptual realism, physical validity, and semantic coherence across both indoor and open-domain scenes.

\subsection{Ablate the Scene Generation Module}
\label{sec:ablate}
\textbf{Setup.} We also conduct an ablation study on \holodeck 2.0's Scene Generation Module. We evaluate two configurations: (1) Using only the initial layouts from the VLM (without the Layout Solver); (2) Using the full Scene Generation Module. For each input text from the Comparative Analysis, we generate 2 scenes (with the same 3D assets) under both settings and then assess their quality using human evaluation. We ask human annotators to choose among 70 pairs of scene layouts, with three options available: ``Prefer A", ``Prefer B",  and ``Tie". Other automated metrics, identical to those in Table \ref{table1}, are also computed.

% \noindent \textbf{Results.} 94 annotators complete the survey, with each annotator answering all 70 questions. Human evaluation shows that the full Scene Generation Module is indispensable for effective 3D scene generation. As the left panel of Figure \ref{fig9} shows, 67.93\% of the choices favor the scenes after our layout, 12.89\% prefer the initial placements generated directly by the VLM, and the remaining 19.18\% are ties.
\medbreak
\noindent \textbf{Results.} A total of 145 annotators complete the survey. Human evaluation shows that the full Scene Generation Module is indispensable for effective 3D scene generation. As shown in Table \ref{table2}, 75.37\% of responses favor our full method for indoor scenes, while 76.30\% favor it for open-domain scenes. Regarding physical plausibility, without the Layout Solver, the \textit{CF} drops to roughly half of that of the full method. Spatial-semantic metrics also decrease significantly: the \textit{PSA} drops from 82.67 to 45.30 for indoor and from 86.00 to 36.53 for open-domain settings.

\begin{figure*}[t]
\centering
\includegraphics[width=1.0\textwidth]{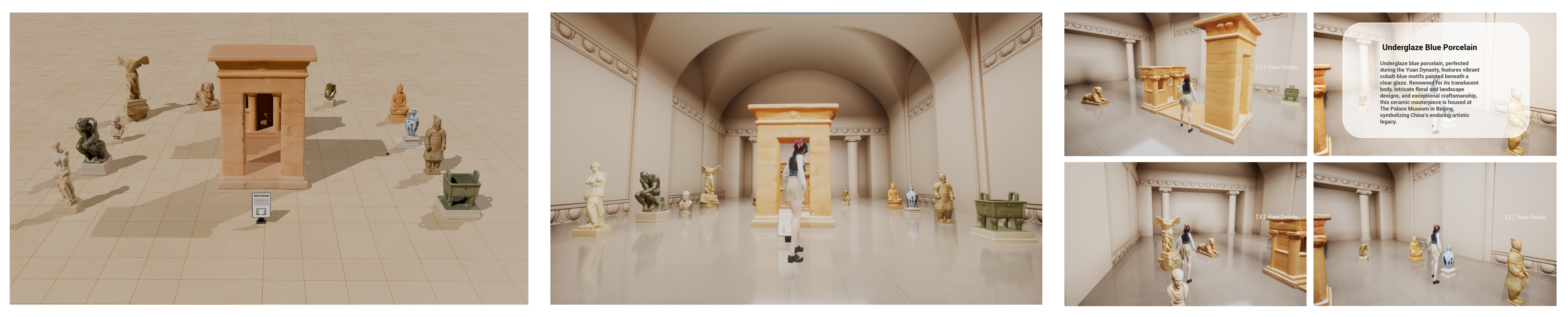}
\caption{3D Scenes generated by \holodeck 2.0 can be seamlessly integrated into standard game production pipelines. From left to right: the \holodeck 2.0 output—a museum hall featuring the Temple of Dendur, the Venus de Milo, The Thinker, etc.; the same scene imported into Unreal Engine and developed into a museum-guide game; and the in-game interface showcasing high-fidelity, fully interactive assets. }
\label{fig6}
\end{figure*}

\begin{figure}[t]
\centering
\includegraphics[width=0.45\textwidth]{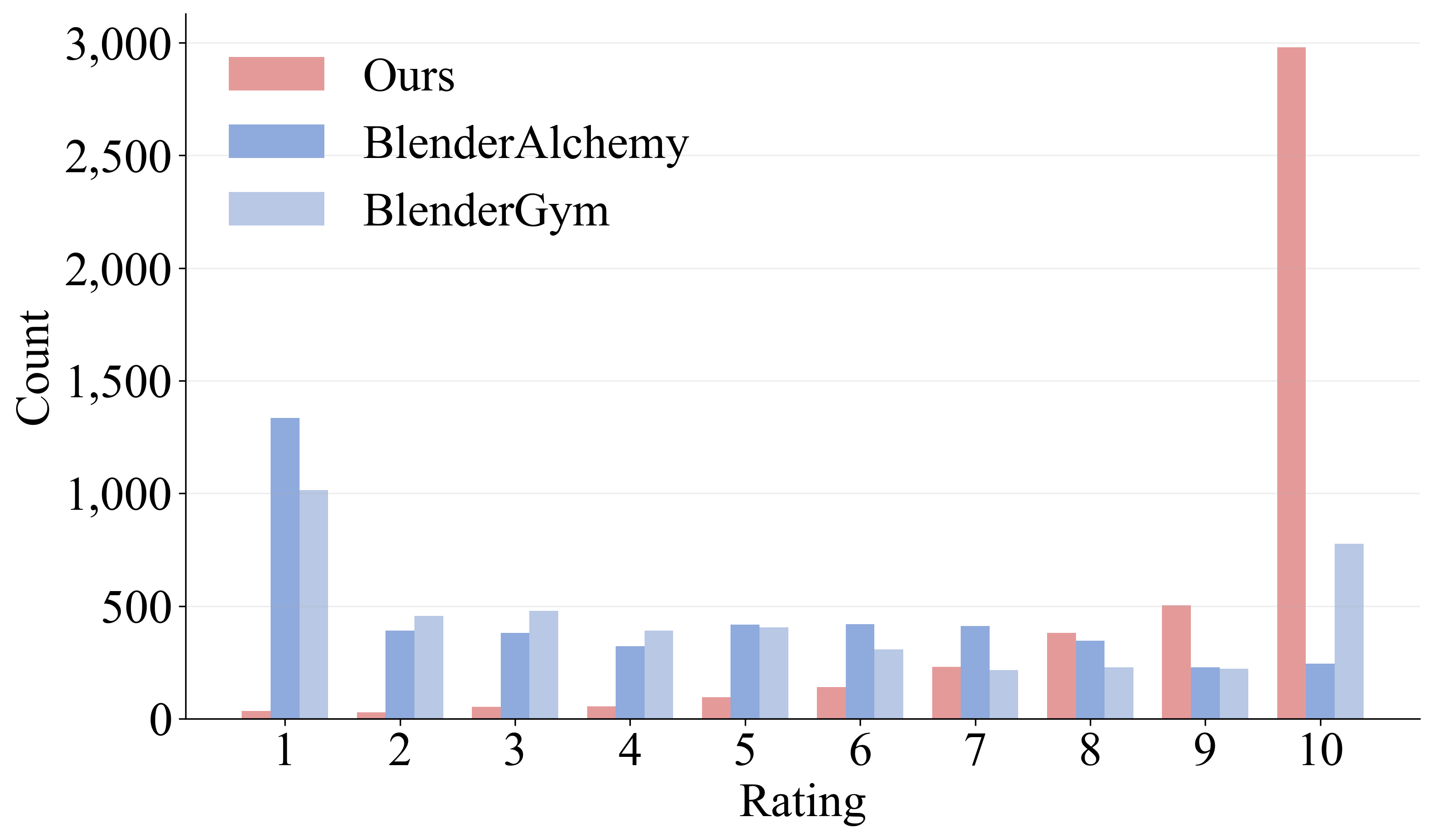} 
\caption{\textbf{Human evaluation results on scene editing.} 
The rating distributions (1–10 scale) of \holodeck 2.0, BlenderAlchemy, and BlenderGym are shown for object placement tasks. }
\label{fig7}
\end{figure}

\subsection{Evaluation on Scene Editing}
\label{sec:edit}
\noindent \textbf{Baselines.} We compare \holodeck 2.0 against two representative 3D scene editing methods: BlenderAlchemy \cite{huang2024blenderalchemy} and BlenderGym \cite{gu2025blendergym}. BlenderAlchemy \cite{huang2024blenderalchemy} leverages VLMs to generate Blender editing sequences by iteratively refining Python programs from user intentions. BlenderGym \cite{gu2025blendergym} is an extension of BlenderAlchemy, with better performance in object placement tasks.
\medbreak
\noindent \textbf{Setup.} We conduct a human evaluation to compare \holodeck 2.0 against both BlenderAlchemy and BlenderGym in placement editing tasks. In our evaluation study: (1) We select 10 representative placement editing tests and apply all three methods to each, producing 30 edited scenes in total. (2) Annotators are then required to rate each edited scene on a scale of 1 to 10 based on overall editing quality and consistency with the instructions.
\medbreak
\noindent \textbf{Results.} 450 annotators complete the survey. As shown in Figure \ref{fig7}, our edited scenes achieve the highest average rating of 9.06, compared to 4.29 for BlenderAlchemy and 4.87 for BlenderGym. This demonstrates that our integrated approach combining DFS and VLM-generated spatial constraints for layout editing outperforms the methods relying solely on VLM-based editing.

In addition to object placement tasks, our method also surpasses these two approaches in style replacement, handling tasks they struggle with, including adding objects in particular artistic styles and replacing existing objects with customized stylistic variations (as shown in Figure \ref{fig5}).

\subsection{Applications in Gaming}
\label{sec:app}
Furthermore, we explore the potential of directly integrating \holodeck 2.0-generated scenes into commercial game engines for game development. To show this, we utilize \holodeck 2.0 to construct a detailed museum exhibition hall based on rich textual descriptions specifying each artifact’s type, material, style, and placement. \holodeck 2.0 produces a 3D museum scene that closely matches the text, featuring the Temple of Dendur at its center, surrounded by over a dozen objects such as the Venus de Milo and a Shang‑dynasty bronze ding cauldron. This generated 3D scene can then be directly imported into professional game engines such as Unreal Engine, enabling quick and intuitive development of interactions between the game character and artifacts. Additionally, physics is enabled to ensure realistic object interactions. The results in Figure \ref{fig6} demonstrate \holodeck 2.0's high-quality, highly faithful scene generation from detailed text, highlighting its potential for direct application in game content creation.

% \subsection{Efficiency Analysis}
% \label{sec:efficiency}
% We evaluate the computational efficiency of \holodeck 2.0 on a single NVIDIA A6000 GPU. For a typical scene with 10--15 objects, the entire generation process takes about one hour, with 3D asset generation (via a locally deployed Hunyuan3D model) contributing roughly 80\% of the runtime. The time scales approximately linearly with the number of objects, but parallelizing asset generation across multiple GPUs reduces total generation time to about 5 minutes.
\section{Conclusions and Discussions}
We propose \holodeck 2.0, a unified vision-language-guided framework that achieves high-fidelity, stylistically consistent, and editable 3D world generation. We validate the effectiveness of \holodeck 2.0 through human and model evaluations. We also show its potential in game modeling. In future work, we intend to improve \holodeck 2.0's capacity for generating more complex and finely detailed 3D worlds, facilitating the creation of large-scale game environments. Additionally, we aim to integrate it with world models, enabling agents to navigate and interact with the generated environments.

{
    \small
    \bibliographystyle{ieeenat_fullname}
    \bibliography{main}
}


\clearpage
\setcounter{page}{1}

\maketitlesupplementary

% \section{Rationale}
% \label{sec:rationale}
% % 
% Having the supplementary compiled together with the main paper means that:
% % 
% \begin{itemize}
% \item The supplementary can back-reference sections of the main paper, for example, we can refer to \cref{sec:intro};
% \item The main paper can forward reference sub-sections within the supplementary explicitly (e.g. referring to a particular experiment); 
% \item When submitted to arXiv, the supplementary will already included at the end of the paper.
% \end{itemize}
% % 
% To split the supplementary pages from the main paper, you can use \href{https://support.apple.com/en-ca/guide/preview/prvw11793/mac#:~:text=Delete%20a%20page%20from%20a,or%20choose%20Edit%20%3E%20Delete).}{Preview (on macOS)}, \href{https://www.adobe.com/acrobat/how-to/delete-pages-from-pdf.html#:~:text=Choose%20%E2%80%9CTools%E2%80%9D%20%3E%20%E2%80%9COrganize,or%20pages%20from%20the%20file.}{Adobe Acrobat} (on all OSs), as well as \href{https://superuser.com/questions/517986/is-it-possible-to-delete-some-pages-of-a-pdf-document}{command line tools}.

\section{Details of \holodeck 2.0}
\subsection{Efficiency Analysis}
\label{sec:efficiency}
We evaluate the computational efficiency of \holodeck 2.0 on a single NVIDIA A6000 GPU. For a typical scene with 10--15 objects, the entire generation process takes about one hour, with 3D asset generation (via a locally deployed Hunyuan3D model) contributing roughly 80\% of the runtime. The time scales approximately linearly with the number of objects, but parallelizing asset generation across multiple GPUs reduces total generation time to about 5 minutes.

\subsection{Details of the Scene Analysis Module}
In the Scene Analysis Module, \holodeck 2.0 first uses GPT-Image-1 to generate a reference image based on the input text. The prompt is shown below.

\noindent\rule{\linewidth}{1pt}
\vspace{-2em}
\captionof{listing}{Prompt - Generate Reference Image}
\vspace{-0.7em}
\noindent\rule{\linewidth}{0.3pt}
\begin{lstlisting}[language=Python]
prompt = (
    f"{description}. Render in {style} style. 3-D view: x->right, y->backward, z->up. "
    "Well-lit, no extra objects."
)
\end{lstlisting}
\vspace{-0.7em}
\noindent\rule{\linewidth}{0.3pt}

Then, \holodeck 2.0 feeds the input text and the reference image into GPT-o3, which outputs a JSON file containing object properties. The prompt is shown below. We also illustrate the structure of the JSON file through an example.

\noindent\rule{\linewidth}{1pt}
\vspace{-2em}
\captionof{listing}{Prompt - Generate Object Properties}
\vspace{-0.7em}
\noindent\rule{\linewidth}{0.3pt}
\begin{lstlisting}[language=Python]
class Vec3(BaseModel):
    x: float
    y: float
    z: float

class SceneItem(BaseModel):
    id: str
    name: str
    position: Vec3
    rotation: Vec3
    size: Vec3
    visual_description: str

class SceneData(BaseModel):
    items: List[SceneItem]

SCHEMA_JSON = SceneData.model_json_schema()


INSTRUCTION = (
    "You will receive a scene description and a style. Your task is to divide the scene into several objects and provide their positions, rotations, sizes, and visual descriptions in JSON format.\n\n"
    "Your output must be only valid JSON matching this schema:\n\n"
    f"{json.dumps(SCHEMA_JSON, indent=2)}\n\n"
    "Please follow these coordinate system specifications:\n"
    "- Use a right-handed coordinate system\n"
    "- The origin (0,0,0) is at the centre of the floor/ground\n"
    "- +X = right, +Y = backward, +Z = up (metres)\n\n"
    "The size values should be in range [0.1, 5] metres.\n\n"
    "Focus on the main objects in the scene, do not include too many small details.\n\n"
    "Do not include background elements (e.g., walls, windows, ceilings, floors, sky, rivers, grass, roads, terrain, curtains) in the items list.\n\n"
    "Each entry in the items list must correspond to a complete object (e.g. a building, a sculpture)--"
    "do not split out parts or components (e.g., handles, pillars, windows, knobs, etc.) as separate objects.\n\n"
    "_No extra commentary or Markdown--just the JSON text._\n\n"
)
\end{lstlisting}
\vspace{-0.7em}
\noindent\rule{\linewidth}{0.3pt}

\noindent\rule{\linewidth}{1pt}
\vspace{-2em}
\captionof{listing}{JSON File — Object Properties} 
\vspace{-0.7em}
\noindent\rule{\linewidth}{0.3pt}
\begin{lstlisting}[language=Python]
{
  "items": [
    {
      "id": "bed1",
      "name": "King Bed",
      "position": {
        "x": 0.0,
        "y": 1.0,
        "z": 0.4
      },
      "rotation": {
        "x": 0.0,
        "y": 0.0,
        "z": 0.0
      },
      "size": {
        "x": 1.92,
        "y": 1.94,
        "z": 1.2
      },
      "visual_description": "A king-sized bed with a light-beige upholstered headboard, dressed in a white duvet and multiple decorative pillows."
    }, 
    {
      "id": "nightstand_right",
      "name": "Right Nightstand",
      "position": {
        "x": 1.2,
        "y": 1.0,
        "z": 0.35
      },
      "rotation": {
        "x": 0.0,
        "y": 0.0,
        "z": 0.0
      },
      "size": {
        "x": 0.52,
        "y": 0.95,
        "z": 0.6
      },
      "visual_description": "A matching light-wood minimalist nightstand with one drawer and open lower shelf."
    },
    {
      "id": "lamp_left",
      "name": "Left Table Lamp",
      "position": {
        "x": -1.2,
        "y": 1.0,
        "z": 0.9
      },
      "rotation": {
        "x": 0.0,
        "y": 0.0,
        "z": 0.0
      },
      "size": {
        "x": 0.34,
        "y": 0.34,
        "z": 0.6
      },
      "visual_description": "A sleek glass-base table lamp with a white drum shade."
    },
    ...
  ]
}
\end{lstlisting}
\vspace{-0.7em}
\noindent\rule{\linewidth}{0.3pt}

Then, \holodeck 2.0 extracts object names from the JSON file to form a list. This list, along with the scene reference image, is then passed to GPT-Image-1 to generate frontal images for individual objects. The prompt is shown below.
\noindent\rule{\linewidth}{1pt}
\vspace{-2em}
\captionof{listing}{Prompt - Generate Individual Images}
\vspace{-0.7em}
\noindent\rule{\linewidth}{0.3pt}
\begin{lstlisting}[language=Python]
prompt = (
    f"Please generate ONE PNG image of an isolated front-view {obj_name} "
    f"with a transparent background, in {style} style, with shapes and style similar to the reference scene. "
)
  \end{lstlisting}
\vspace{-0.7em}
\noindent\rule{\linewidth}{0.3pt}

After generating the individual images for different objects, \holodeck 2.0 uses GPT-o3 again for quality control. For example, sometimes a generated image of a large object may already include a smaller object (e.g., a bathtub image containing a faucet), while a separate image of the smaller object is also generated. \holodeck 2.0 prompts GPT-o3 to correct such cases by removing redundant images of the smaller objects. The prompt is shown below.

\noindent\rule{\linewidth}{1pt}
\vspace{-2em}
\captionof{listing}{Prompt - Remove Redundant Images}
\vspace{-0.7em}
\noindent\rule{\linewidth}{0.3pt}
\begin{lstlisting}[language=Python]
system_msg = {
    "role": "system",
    "content": "You are a helpful assistant that identifies redundant sub-components in a set of images."
}

class ImageList(BaseModel):
    filenames: list[str]
SCHEMA_JSON = ImageList.model_json_schema()

instruction = (
    f"""
    I have the following list of image filenames:
    {json.dumps(png_files, indent=2)}
    """
    "Here are all images in the current scene. Some represent smaller parts already "
    "contained within larger assemblies. Return a JSON list of filenames (with ".png") "
    "that should be deleted as redundant sub-components, matching this schema: \n"
    f"{json.dumps(SCHEMA_JSON, indent=2)}\n\n"
)
\end{lstlisting}
\vspace{-0.7em}
\noindent\rule{\linewidth}{0.3pt}

To better capture background information, \holodeck 2.0 feeds the scene reference image into GPT-Image-1 to extract a background reference image, with a primary focus on the ground background. We aim for the generated images to have a uniform ground surface. The prompt is shown below.

\noindent\rule{\linewidth}{1pt}
\vspace{-2em}
\captionof{listing}{Prompt - Generate Background Image} 
\vspace{-0.7em}
\noindent\rule{\linewidth}{0.3pt}
\begin{lstlisting}[language=Python]
prompt = (
    "Replace the entire image with ONE seamless, tileable PNG of the main floor for indoor scenes, and the main ground for outdoor scenes. "
    "Using the material and pattern seen in the input photo. Ignore walls, ceiling and decorations."
    "The texture must be homogeneous, repeating smoothly, and produced at a scale large enough to cover an expansive floor area. Do not add transparency."
)
\end{lstlisting}
\vspace{-0.7em}
\noindent\rule{\linewidth}{0.3pt}

\subsection{Details of the Object Generation Module}

In the Object Generation Module, we run the deployed Hunyuan3D 2.1 model locally. Given the individual image of each object, Hunyuan3D 2.1 generates a corresponding 3D asset in GLB-format. After generating the 3D assets for each object, we write the assets' file path back into the JSON file containing the object properties. We also rescale each 3D asset based on its actual x, y, and z dimensions, using the original z size as a reference. The updated x and y sizes after scaling are then written back into the JSON file.

\subsection{Details of the Scene Generation Module}
In the Scene Generation Module, \holodeck 2.0 first embeds the scene input text along with each object's ID, name, and visual description (extracted from the JSON file) into the prompt. This prompt, together with the scene reference image, is then fed into GPT-o3 to generate the necessary spatial constraints between objects in the scene. Examples of such spatial constraints are provided below.

\noindent\rule{\linewidth}{1pt}
\vspace{-2em}
\captionof{listing}{JSON File - Spatial Constraints}
\vspace{-0.7em}
\noindent\rule{\linewidth}{0.3pt}
\begin{lstlisting}[language=Python]
[
  {
    "type": "relative",
    "relation": "left of",
    "source": "nightstand_left",
    "target": "bed1"
  },
  {
    "type": "relative",
    "relation": "right of",
    "source": "nightstand_right",
    "target": "bed1"
  },
  {
    "type": "relative",
    "relation": "right of",
    "source": "dresser1",
    "target": "bed1"
  },
  {
    "type": "relative",
    "relation": "on",
    "source": "lamp_left",
    "target": "nightstand_left"
  },
  {
    "type": "distance",
    "relation": "near",
    "source": "lamp_left",
    "target": "bed1"
  },
  {
    "type": "relative",
    "relation": "on",
    "source": "lamp_right",
    "target": "nightstand_right"
  },
  {
    "type": "distance",
    "relation": "near",
    "source": "lamp_right",
    "target": "bed1"
  },
  {
    "type": "relative",
    "relation": "on",
    "source": "photo_frames1",
    "target": "dresser1"
  }
]
\end{lstlisting}
\vspace{-0.7em}
\noindent\rule{\linewidth}{0.3pt}

It is worth noting that, to facilitate the operation of our subsequent DFS solver, we specify to GPT-o3 that once an object has been used as a target, it can no longer be used as a source in any following constraints. We also require that constraints with the same source must appear consecutively. The prompt is shown below.

\noindent\rule{\linewidth}{1pt}
\vspace{-2em}
\captionof{listing}{Prompt - Generate Spatial Constraints}
\vspace{-0.7em}
\noindent\rule{\linewidth}{0.3pt}
\begin{lstlisting}[language=Python]
system_prompt = (
    "You are a spatial relationship analyzer. Generate valid JSON constraints based on scene descriptions, object information and the reference image."
    "Output the constraints in a strict sequence: once an object has appeared as a target, it must not later appear as a source."
    "For each object, output all of its source-type constraints in true consecutive order in the list - they must be physically adjacent without any other constraints interleaved.")

user_prompt = f"""
You are a spatial relationship analyzer. Given a scene description and a list of objects with their IDs, generate spatial constraints between the objects.

Scene Description:
{description}

Available Objects:
{objects_text}

Generate spatial constraints in the following JSON format:
[
  {{
    "type": "relative",
    "relation": "right of|left of|in front of|behind|side of|on|above",
    "source": "object_id",
    "target": "object_id"
  }},
  {{
    "type": "distance", 
    "relation": "near|far",
    "source": "object_id",
    "target": "object_id"
  }},
  {{
    "type": "rotation",
    "relation": "face to",
    "source": "object_id", 
    "target": "object_id"
  }}
]

Guidelines:
1. Only include meaningful spatial relationships that match the description
2. Use the exact object IDs from the available objects list
3. Focus on the most important spatial relationships
4. Avoid redundant relationships
5. Return only valid JSON, no additional text
6. For example, if the scene description mentions "a chair to the right of a table", the relation should be right of, with the chair as the source and the table as the target.
7. An object should either choose to be on the ground(default, no need to specify as a constraint) or on/above another object.
8. Use "on" when one object is physically resting on the surface of another object. Use "above" only when the object is floating or suspended in the air above another object, not touching it. If an object is on the ground, do not specify any "on" or "above" relation when it is the source.

Generate the constraints:
"""
\end{lstlisting}
\vspace{-0.7em}
\noindent\rule{\linewidth}{0.3pt}

Next, we input the spatial constraints along with the JSON file containing object properties into the DFS solver. The DFS solver works by placing objects one by one based on the dependency structure defined by the constraints. Objects that do not depend on any others are placed first, preferably at their initial positions inferred by the large reasoning model GPT-o3. Subsequently, other objects are placed in order of their dependencies — each object is only placed once all its constraints are satisfied and no collisions are detected. The solver proceeds with this search process until all objects are successfully placed, at which point it outputs the final layout.

If some objects cannot be placed due to conflicting constraints, the DFS solver will report which object's constraints could not be fully satisfied during the search. We then use this information as part of the prompt (edit\_instructions) to GPT-o3 to regenerate the spatial constraints, allowing us to iteratively refine the layout. The prompt used for regenerating the constraints is shown below.

\noindent\rule{\linewidth}{1pt}
\vspace{-2em}
\captionof{listing}{Prompt - Regenerate Spatial Constraints}
\vspace{-0.7em}
\noindent\rule{\linewidth}{0.3pt}
\begin{lstlisting}[language=Python]
prompt = f"""
You are a spatial relationship analyzer. Given a scene description and a list of objects with their IDs, generate spatial constraints between the objects.

Scene Description:
{description}

Available Objects:
{objects_text}

Last time, the following constraints were generated:
{last_constraint_content}

Edit Instructions:
{edit_instructions}

Generate spatial constraints in the following JSON format:
[
  {{
    "type": "relative",
    "relation": "right of|left of|in front of|behind|side of|on|above",
    "source": "object_id",
    "target": "object_id"
  }},
  {{
    "type": "distance", 
    "relation": "near|far",
    "source": "object_id",
    "target": "object_id"
  }},
  {{
    "type": "rotation",
    "relation": "face to",
    "source": "object_id", 
    "target": "object_id"
  }}
]

Guidelines:
1. Only include meaningful spatial relationships that match the description
2. Use the exact object IDs from the available objects list
3. Focus on the most important spatial relationships
4. Avoid redundant relationships
5. Return only valid JSON, no additional text
6. For example, if the scene description mentions "a chair to the right of a table", the relation should be right of, with the chair as the source and the table as the target.
7. An object should either choose to be on the ground(default, no need to specify as a constraint) or on/above another object.
8. Use "on" when one object is physically resting on the surface of another object. Use "above" only when the object is floating or suspended in the air above another object, not touching it. If an object is on the ground, do not specify any "on" or "above" relation when it is the source.

Generate the constraints:
"""
\end{lstlisting}
\vspace{-0.7em}
\noindent\rule{\linewidth}{0.3pt}

\subsubsection{Spatial Constraints Definition}
Every statement extracted from the language prompt is normalized to a \emph{constraint}
\[
c=\langle\textit{type},\;\textit{relation},\;\textit{source},\;\textit{target}\rangle ,
\]
where \textit{source} and \textit{target} are object identifiers that already exist in the scene JSON.  
The recognized categories and their semantics are summarized in Table 3.
\begin{table*}[t]
  \centering
  \renewcommand{\arraystretch}{1.2}
  \begin{tabular}{@{} l l p{8.5cm} @{}}
    \toprule
    \textbf{Type} & \textbf{Relation} & \textbf{Geometric Meaning} \\
    \midrule
    \textit{relative} & \textit{left of}      &
      Planar ordering along the $x$–axis: the source lies to the left of (negative $x$) the target’s bounding box, with a 0.1\,m buffer. \\
    \addlinespace
                      & \textit{right of}     &
      Planar ordering along the $x$–axis: the source lies to the right of (positive $x$) the target’s bounding box, with a 0.1\,m buffer. \\
    \addlinespace
                      & \textit{in front of}  &
      Planar ordering along the $y$–axis: the source lies in front of (negative $y$) the target’s bounding box, with a 0.1\,m buffer. \\
    \addlinespace
                      & \textit{behind}       &
      Planar ordering along the $y$–axis: the source lies behind (positive $y$) the target’s bounding box, with a 0.1\,m buffer. \\
    \addlinespace
                      & \textit{side of}      &
      Planar ordering on the $x$–axis: the source lies to either lateral side of the target beyond its bounding box, with a 0.1\,m buffer. \\
    \midrule
    \textit{distance} & \textit{near}         &
      The horizontal gap between source and target is at most 2\,m. \\
    \addlinespace
                      & \textit{far}          &
      The horizontal gap between source and target exceeds 8\,m. \\
    \midrule
    \textit{vertical} & \textit{on}           &
      The source rests directly on the target’s top surface, with clearance under 2\,mm. \\
    \addlinespace
                      & \textit{above}        &
      The source is positioned at least 2\,m above the target, while still overlapping its footprint. \\
    \midrule
    \textit{rotation} & \textit{face to}      &
      The source’s forward-facing direction points at the target’s center within ±10°. \\
    \bottomrule
  \end{tabular}
  \caption*{Table 3: Constraint primitives used by the Scene Generation Module.}
\end{table*}

Before the search phase we apply two global authoring rules:

\begin{itemize}
  \item \textbf{Ground-only default.}  An object either \emph{rests on the ground} or is explicitly
        the \textit{source} of a vertical relation — never both.
  \item \textbf{Acyclic dependency.}  When generating constraints, once an object has appeared as a \textit{target} it cannot later re-appear as a \textit{source}, yielding a partial order that the solver exploits.
\end{itemize}

All constraints are collected into a flat JSON list with duplicate
and self-referential entries removed.
\subsubsection{DFS Solver Implementation}
The DFS‐based solver proceeds as follows:

\begin{itemize}
  \item \textbf{Preprocessing.}  We first build a lookup table of each object’s size, initial pose, and applicable constraints, then perform a topological sort of the objects according to the acyclic dependency rule (any object that appears as a target may not later serve as a source).  
  \item \textbf{Candidate Generation.}  For each object in sorted order, we collect its active constraints against already‐placed objects.  
    \begin{itemize}
      \item If it has no constraints, we sample positions in a small neighbourhood around its initial location.  
      \item If it has exactly one constraint, we invoke the relation‐specific generator.  
      \item If it has multiple constraints, we generate candidates from the highest‐priority constraint and then filter out any that violate the others.  
    \end{itemize}
  \item \textbf{Collision and Constraint Checking.}  Each candidate placement is immediately tested for mesh‐level collisions and for precise satisfaction of all its constraints. Invalid placements are discarded before recursion.
  \item \textbf{Recursive Search and Backtracking.}  We place the object, recurse to the next one, and upon return remove it if we need to explore other branches.  The search is depth‐first, but we track a global node counter and elapsed time: if either limit is reached, we abort further recursion and return the best complete (or partial) layout found so far.
  \item \textbf{Solution Selection.}  Whenever all objects have been placed, we compare the solution’s score (number of successfully placed items) against our current best.  The highest‐scoring layout is stored and ultimately returned when the search finishes or is cut short.
\end{itemize}
% \begin{algorithm}[tb]
% \caption{Example algorithm}
% \label{alg:algorithm}
% \textbf{Input}: Your algorithm's input\\
% \textbf{Parameter}: Optional list of parameters\\
% \textbf{Output}: Your algorithm's output
% \begin{algorithmic}[1] %[1] enables line numbers
% \STATE Let $t=0$.
% \WHILE{condition}
% \STATE Do some action.
% \IF {conditional}
% \STATE Perform task A.
% \ELSE
% \STATE Perform task B.
% \ENDIF
% \ENDWHILE
% \STATE \textbf{return} solution
% \end{algorithmic}
% \end{algorithm}

% \begin{algorithm}[t]
% \caption{DFS-Based Layout Solver}
% \label{alg:dfs_solver}
\noindent\rule{\linewidth}{1pt}
\vspace{-2em}
\captionof{algorithm}{DFS-Based Layout Solver}\label{alg:dfs_solver}
\vspace{-0.7em}
\noindent\rule{\linewidth}{0.3pt}
\textbf{Input}: Set of items $I$, spatial constraints $C$\\
\textbf{Parameter}: Timeout $T$, node limit $N$\\
\textbf{Output}: Best placement $\mathrm{best\_sol}$
\begin{algorithmic}[1]
  \STATE Build map of each item’s size, initial position, and rotation
  \STATE Topologically sort $I$ into ordered list $O$ (respecting acyclic dependencies)
  \STATE $\mathrm{best\_score}\leftarrow -\infty$, $\mathrm{best\_sol}\leftarrow\emptyset$
  \STATE Record $\mathrm{start\_time}\leftarrow$ now, $\mathrm{node\_cnt}\leftarrow 0$
  \STATE \textbf{procedure} DFS($i$, placed)
    \IF{$i = |O|$}
      \STATE Update $\mathrm{best\_sol}$ if current score $> \mathrm{best\_score}$
      \STATE \textbf{return}
    \ENDIF
    \STATE $x \leftarrow O[i]$
    \STATE Collect all constraints in $C$ with source $=x$ and target $\in$ placed
    \IF{no such constraints}
      \STATE Generate candidate positions around $x$’s initial location
    \ELSE
      \IF{exactly one constraint}
        \STATE Generate candidate positions according to that single relation
      \ELSE
        \STATE Generate candidate positions by the primary constraint, then filter by the rest
      \ENDIF
    \ENDIF
    \FOR{each candidate in candidates}
      \IF{collision detected or any constraint check fails}
        \STATE \textbf{continue}
      \ENDIF
      \STATE placed[$x$] $\leftarrow$ candidate
      \STATE DFS($i+1$, placed)
      \STATE remove $x$ from placed
      \IF{timeout reached or $\mathrm{node\_cnt} > N$}
        \STATE \textbf{return}
      \ENDIF
    \ENDFOR
    \STATE \textbf{return}
  \STATE \textbf{end procedure}
  \STATE DFS($0$, $\{\}$)
  \STATE \textbf{return} $\mathrm{best\_sol}$
\end{algorithmic}
\vspace{-0.7em}
\noindent\rule{\linewidth}{0.3pt}

\subsection{Details of the Scene Editing Module}
After a scene is generated, users can directly edit the scene through natural language. If a user wants to adjust the spatial relationship of a specific object, they can issue a language command, which will be incorporated into the prompt as an edit instruction. GPT-o3 then uses this prompt to identify the corresponding object from the JSON file and generate new spatial constraints for it. The DFS solver treats all other objects as already placed and searches for a new position for the edited object that satisfies the updated constraints. The prompt used for this process is shown below.

\noindent\rule{\linewidth}{1pt}
\vspace{-2em}
\captionof{listing}{Prompt - Scene Editing Module}
\vspace{-0.7em}
\noindent\rule{\linewidth}{0.3pt}
\begin{lstlisting}[language=Python]
prompt = f"""
You are a spatial relationship analyzer. Given a list of objects with their IDs and a human feedback text, extract spatial constraints.

Your first task is to identify exactly one focus object from the feedback (the only object to be modified). Map this focus object to its exact ID from the Available Objects list. 


Available Objects:
{objects_text}

Human Feedback:
{feedback}

Generate spatial constraints in the following JSON format:
[
  {{
    "type": "relative",
    "relation": "right of|left of|in front of|behind|side of|on|above",
    "source": "object_id",
    "target": "object_id"
  }},
  {{
    "type": "distance", 
    "relation": "near|far",
    "source": "object_id",
    "target": "object_id"
  }},
  {{
    "type": "rotation",
    "relation": "face to",
    "source": "object_id", 
    "target": "object_id"
  }}
]

Guidelines:
1. Identify exactly one focus object from the feedback; it must be the only "source" in all constraints. Never use the focus object as a "target".
2. Generate constraints only if they are explicitly supported by the feedback.
3. Use the exact object IDs from the Available Objects list. Ignore unknown objects.
4. Avoid redundant or contradictory relationships. If conflicts arise, choose the most specific interpretation.
5. Example: if the feedback says "Put the chair to the right of the table", use relation "right of", with the chair as the source and the table as the target.
6. Use "on" only when an object is physically resting on the surface of another object. Use "above" only when the object is suspended or floating above another object without contact. If an object is simply on the ground, do not produce "on" or "above" with it as the source.
7. Return only valid JSON, no additional text

Generate the constraints:
"""
\end{lstlisting}
\vspace{-0.7em}
\noindent\rule{\linewidth}{0.3pt}

If the user wants to replace an object, the Scene Editing Module automatically calls the 3D asset generation model to regenerate the corresponding object and replaces the original with the new one. If the user wants to add an object, the Scene Editing Module automatically calls the 3D asset generation model to generate the new object, then creates spatial constraints for it and uses the DFS solver to find a suitable placement. As for deleting objects, since all objects in the scene are manipulable mesh assets, this operation is trivial.

\subsection{Rendering Options}
To achieve better rendering quality, we choose to perform the rendering in Blender. We adjust the camera among three candidate viewpoints—an overhead isometric view from the left, a frontal view, and a right‐side view—and select the one that best showcases the entire scene. The specific parameters are detailed below:

\vspace{0.3em}
All renders use a resolution of 3840×2160 px, with 1024 samples per pixel under the Cycles engine. GPU acceleration is enabled to speed up sampling, and Blender’s built-in denoising is turned on to reduce noise while preserving fine details.

\subsection{Gaming}
We import the generated GLB-format scene file into Unreal Engine 5.6 for game modeling. We present additional in-game images in Figure \ref{fig8} to demonstrate the richness, high quality, and interactivity of the scenes generated by \holodeck 2.0.

\section{Details of Evaluation}

\paragraph{Notes on Baseline Coverage and Metric Availability.}
When comparing \holodeck 2.0 with existing baselines, several practical constraints affect the completeness of certain metrics and the number of evaluated scenes. 

\vspace{0.3em}
First, GraphDreamer adopts an implicit 3D representation based on Gaussian splatting, where objects are not separable into distinct geometric units. Consequently, collision-based physical metrics (\textit{CF}, \textit{CN}, and \textit{PSA}) cannot be computed, as they require explicit per-object geometry to detect intersections and evaluate physically grounded spatial consistency. In addition, the rendered views of GraphDreamer contain substantial visual clutter, making them unsuitable for human perceptual evaluation, which relies on clearly distinguishable objects and scene structure.

\vspace{0.3em}
Second, GraphDreamer and Scene Language do not yield valid outputs for all scene descriptions in our benchmark. Among the 70 test scenes, GraphDreamer successfully produces only 14 scenes and Scene Language produces 44 scenes, whereas the other baselines and our method generate all 70. For fairness, we include all valid outputs returned by each method without imputing missing results.

\vspace{0.3em}
These factors lead to different numbers of evaluated samples across baselines and explain why certain metrics are unavailable for methods whose representations or outputs do not permit their computation.

\subsection{Human Evaluation}
% We conduct a total of four online questionnaires; the first two focus on static scene generation in indoor and outdoor environments, respectively.  Each of these two questionnaires comprises 35 distinct scenarios.  For each scenario, participants answer eight separate rating questions on a 10-point Likert scale (1 = worst, 10 = best), measuring:
% \begin{itemize}
%   \item Object‐level quality of \holodeck v.s. \holodeck 2.0
%   \item Scene‐level quality of \holodeck v.s. \holodeck 2.0
% \end{itemize}
We conduct a total of four online questionnaires.
\subsubsection{Comparative Analysis}
The first two focus on scene generation quality in indoor and outdoor environments, respectively. Each questionnaire contains 35 distinct scenarios. For every scenario, participants answer eight rating questions on a 1--10 scale (1 = worst, 10 = best), evaluating:

\begin{itemize}
    \item Object-level quality across \holodeck, PartCrafter, Scene Language, and \holodeck 2.0
    \item Scene-level quality across the above methods
\end{itemize}

These human-rated perceptual scores provide complementary evaluation to automated physical and spatial-semantic metrics, enabling a comprehensive comparison of all baselines against \holodeck 2.0.

Two example items from these questionnaires are shown below:

\vspace{1em}
\textbf{Text:} ``A squat rack stands on rubber flooring, with two vertical weight plate holders attached to its sides. A flat bench is placed parallel in front of the rack at a distance of 1.2 meters. To the right of the rack, a horizontal dumbbell rack, 2 meters wide, holds dumbbells arranged from light to heavy." \\
The rendered 3D scenes generated from this text are shown below.
\vspace{10pt} 
\begin{figure}[H]
  \centering
  \begin{subfigure}[b]{0.45\linewidth}
    \includegraphics[width=\linewidth]{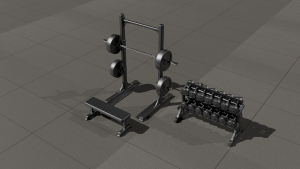}
    \caption{Scene A}
  \end{subfigure}
  \hfill
  \begin{subfigure}[b]{0.45\linewidth}
    \includegraphics[width=\linewidth]{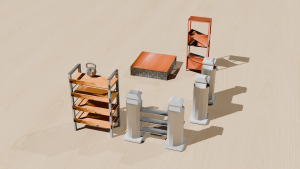}
    \caption{Scene B}
  \end{subfigure}
 \vskip\baselineskip 

  \begin{subfigure}[b]{0.45\linewidth}
    \includegraphics[width=\linewidth]{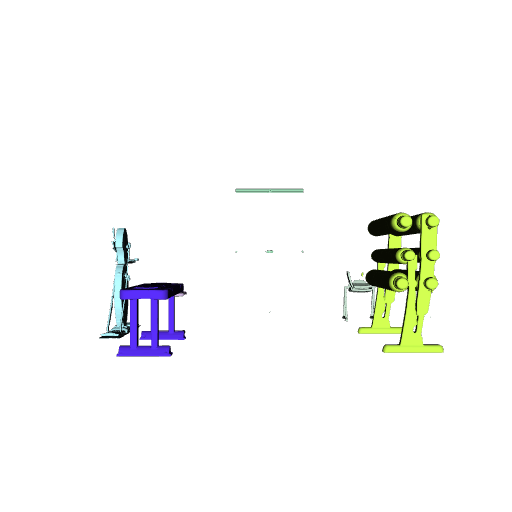}
    \caption{Scene C}
  \end{subfigure}
  \hfill
  \begin{subfigure}[b]{0.45\linewidth}
    \includegraphics[width=\linewidth]{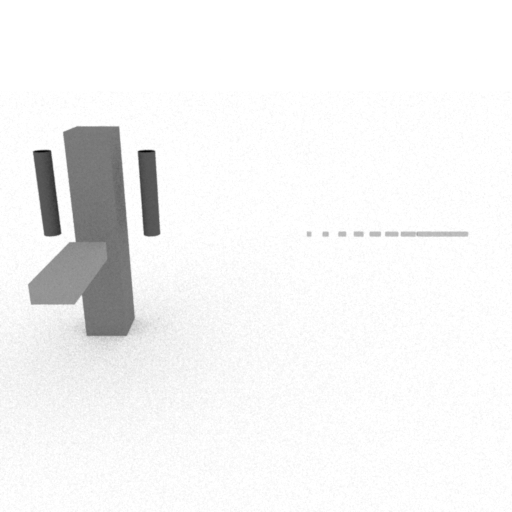}
    \caption{Scene D}
  \end{subfigure}
  \label{fig:case1_examples}
\end{figure}
\begin{enumerate}
    \item Please rate how well the \textbf{objects} generated in Scene A match the text input.
    \item Please rate how well the \textbf{overall scene} in Scene A matches the text input.
    \item Please rate how well the \textbf{objects} generated in Scene B match the text input.
    \item Please rate how well the \textbf{overall scene} in Scene B matches the text input.
    \item Please rate how well the \textbf{objects} generated in Scene C match the text input.
    \item Please rate how well the \textbf{overall scene} in Scene C matches the text input.
    \item Please rate how well the \textbf{objects} generated in Scene D match the text input.
    \item Please rate how well the \textbf{overall scene} in Scene D matches the text input.
\end{enumerate}

\textbf{Text:} ``Realistic Style. A rural roadside scene includes a wooden fruit stand stocked with crates. A pickup truck is parked beside it. A folding table holds jars of jam, and a water barrel rests behind the stand on gravel ground." \\
The rendered 3D scenes generated from this text are shown below.
\begin{figure}[H]
  \centering
  \begin{subfigure}[b]{0.45\linewidth}
    \includegraphics[width=\linewidth]{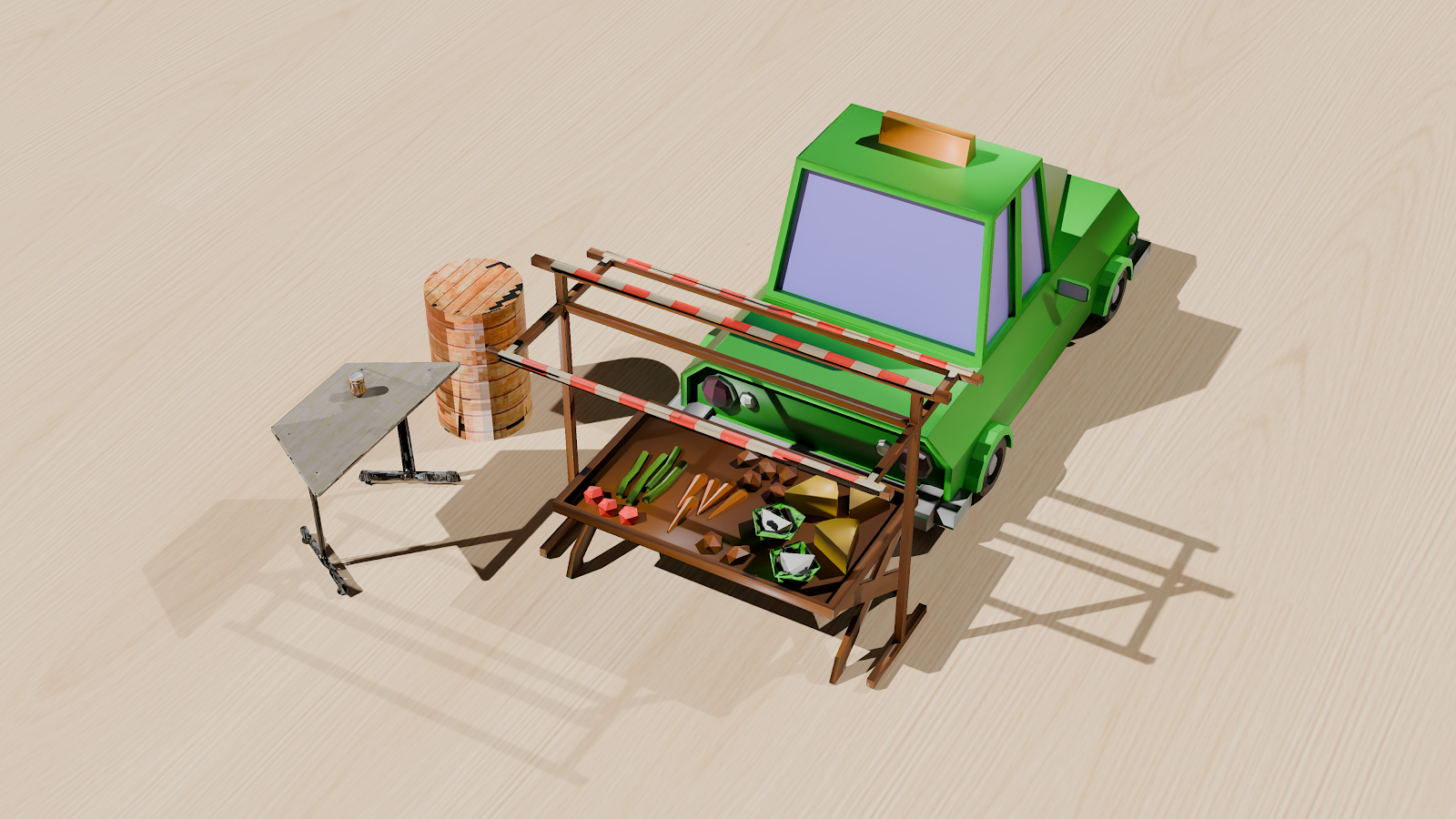}
    \caption{Scene A}
  \end{subfigure}
  \hfill
  \begin{subfigure}[b]{0.45\linewidth}
    \includegraphics[width=\linewidth]{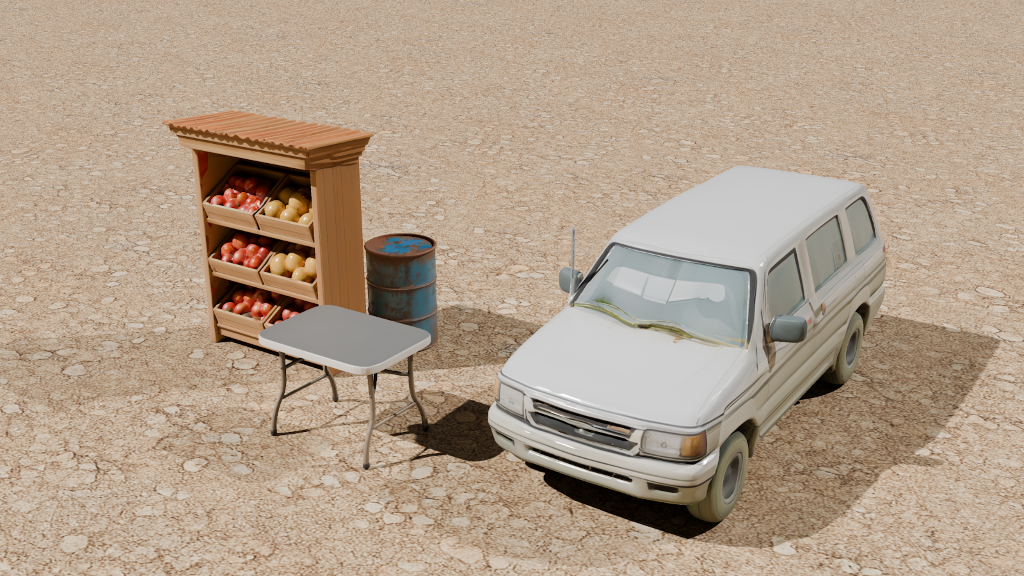}
    \caption{Scene B}
  \end{subfigure}
   \vskip\baselineskip 

  \begin{subfigure}[b]{0.45\linewidth}
    \includegraphics[width=\linewidth]{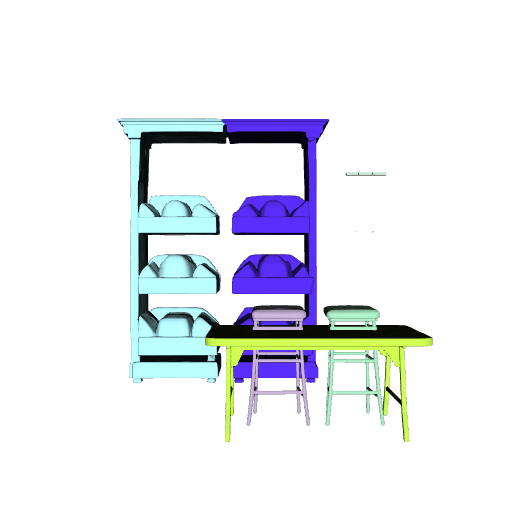}
    \caption{Scene C}
  \end{subfigure}
  \hfill
  \begin{subfigure}[b]{0.45\linewidth}
    \includegraphics[width=\linewidth]{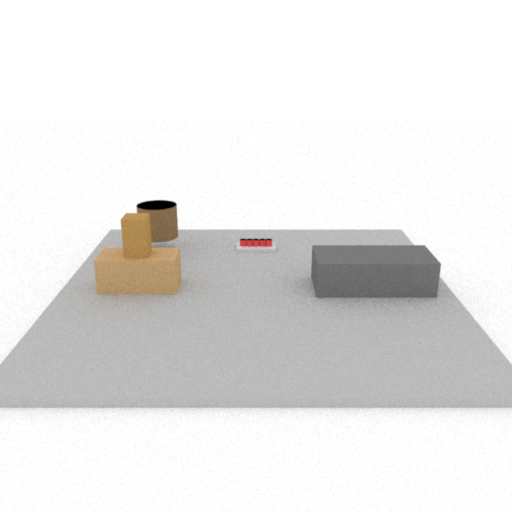}
    \caption{Scene D}
  \end{subfigure}
  % \caption{Generated results for Example Item 2.}
  \label{fig:case2_examples}
\end{figure}

\begin{enumerate}
    \item Please rate how well the \textbf{objects} generated in Scene A match the text input.
    \item Please rate how well the \textbf{overall scene} in Scene A matches the text input.
    \item Please rate how well the \textbf{objects} generated in Scene B match the text input.
    \item Please rate how well the \textbf{overall scene} in Scene B matches the text input.
    \item Please rate how well the \textbf{objects} generated in Scene C match the text input.
    \item Please rate how well the \textbf{overall scene} in Scene C matches the text input.
    \item Please rate how well the \textbf{objects} generated in Scene D match the text input.
    \item Please rate how well the \textbf{overall scene} in Scene D matches the text input.
\end{enumerate}

\subsubsection{Ablate the Scene Generation Module}
The third questionnaire focuses on the ablation study and contains a total of 60 questions. In each question, participants are asked to evaluate the similarity between two rendered images(one for \holodeck 2.0, one for using only the initial layout inferred by GPT-o3) and a given text, and to answer a single-choice question. The options for each question are: ``Prefer A," ``Prefer B," and ``Tie."

An example related to the ablation study is shown below. \vspace{1em}

\textbf{Text:} ``Realistic Style. A winding cobblestone street intersects a small plaza dotted with stone planters. Wooden benches circle a low stone fountain. A row of trimmed hedges lines one side, while scattered fallen leaves drift across the ground. A lone bicycle leans against a lamppost." \\
The two rendered 3D scenes generated from this text are shown below.
\begin{figure}[H]
  \centering
  \begin{subfigure}[b]{0.45\linewidth}
    \includegraphics[width=\linewidth]{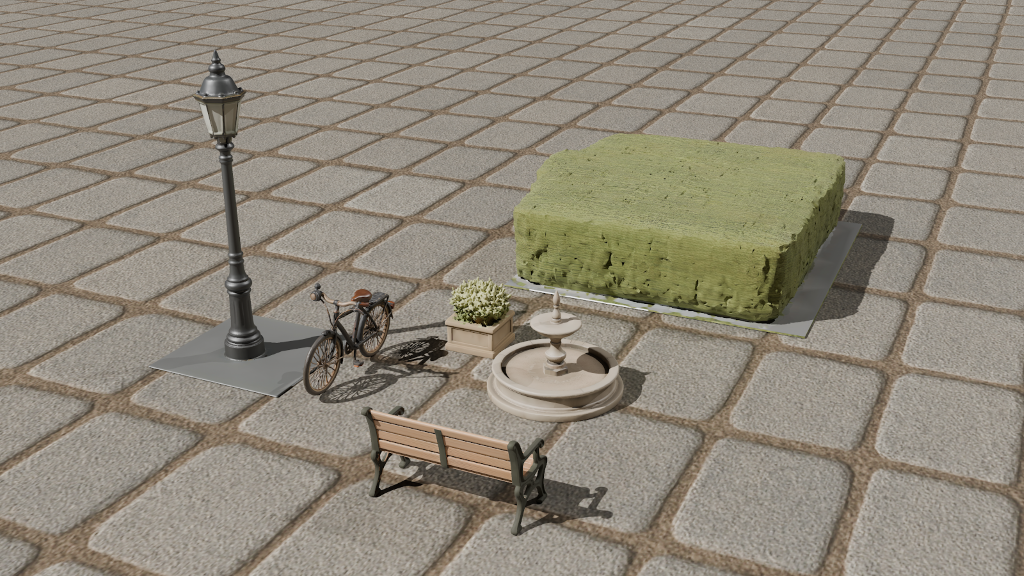}
    \caption{Scene A}
  \end{subfigure}
  \hfill
  \begin{subfigure}[b]{0.45\linewidth}
    \includegraphics[width=\linewidth]{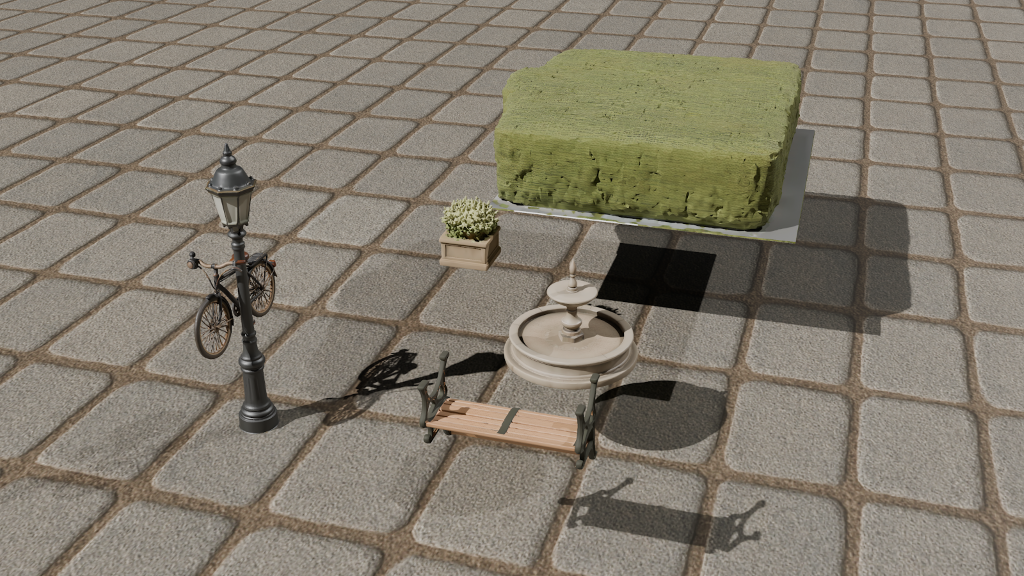}
    \caption{Scene B}
  \end{subfigure}
  % \caption{Generated results for Example Item 3.}
  \label{fig:case3_examples}
\end{figure}

Which scene do you prefer?

\begin{enumerate}
    \item Prefer A
    \item Prefer B
    \item Tie
\end{enumerate}

\subsubsection{Evaluation on Scene Editing}
The fourth questionnaire focuses on the comparison analysis of scene editing tasks. We perform 10 representative placement editing tests for BlenderAlchemy, BlenderGym and \holodeck 2.0.  Annotators are then required to choose their preferred version between the two edited scenes. An example related to the comparison analysis is shown below.

\vspace{1em}

The original scene is shown below:
\begin{figure}[H]
  \centering
    % \vspace{-20pt}
    \includegraphics[width=0.8\linewidth]{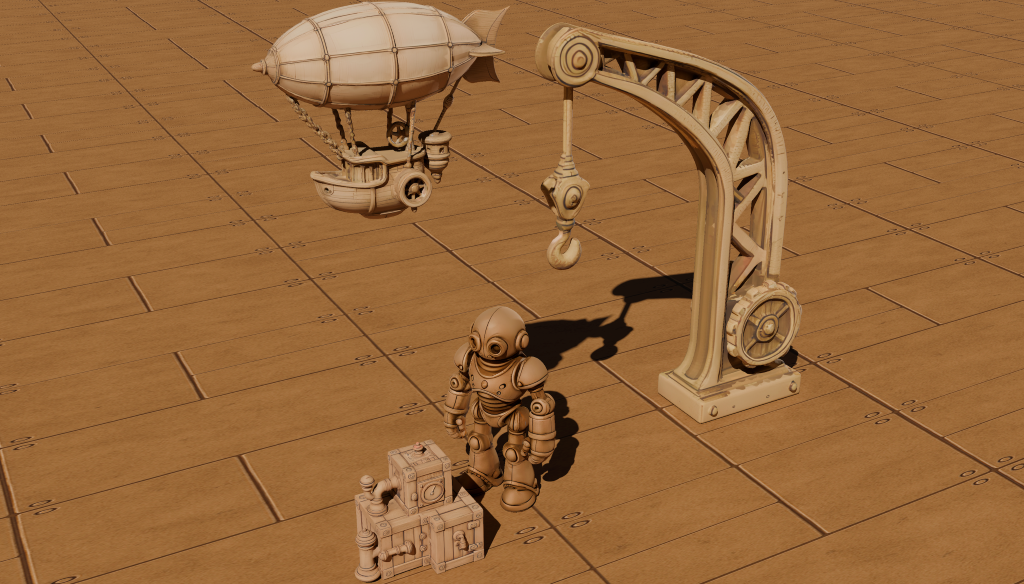}
    \caption*{Original Scene}
    \vspace{-10pt}
\end{figure}

Please determine which of the following edited 3D scenes best matches the editing instruction: ``Move the robot to the top of the stack, facing to the crane."

\begin{figure}[H]
  \centering
    % \vspace{-20pt}
    \includegraphics[width=0.8\linewidth]{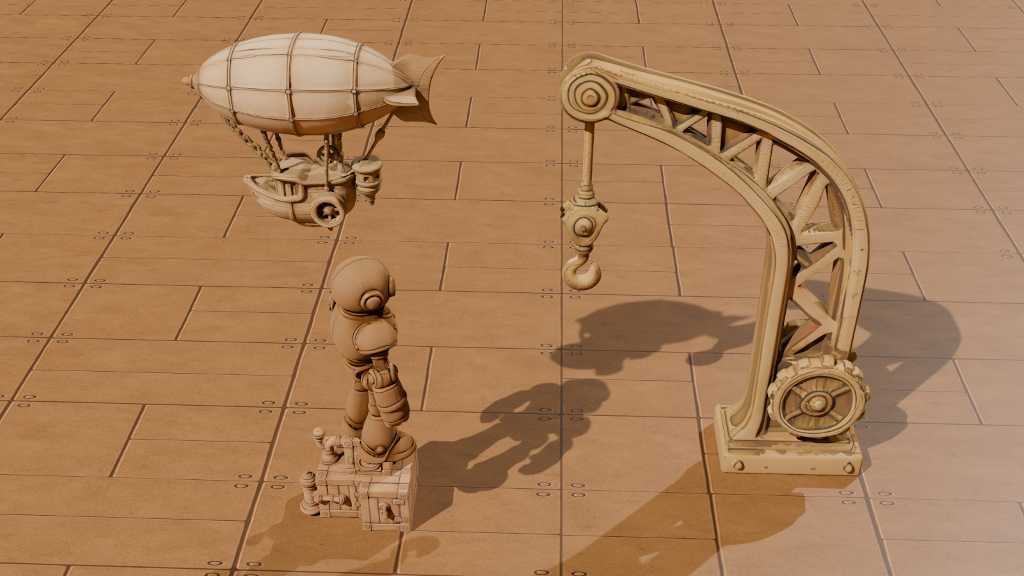}
    \caption*{Scene A}
    \vspace{-15pt}
\end{figure}
\begin{enumerate}
    \item How would you rate the quality of this edited result? (From 1 to 10)
\begin{figure}[H]
  \centering
    % \vspace{-20pt}
    \includegraphics[width=0.8\linewidth]{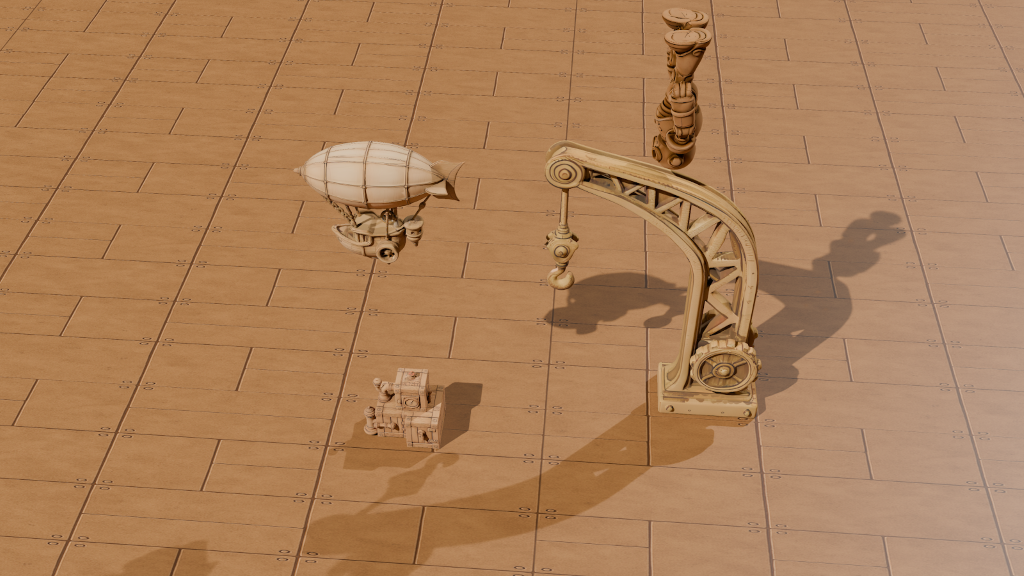}
    \caption*{Scene B}
    \vspace{-15pt}
\end{figure}
    \item How would you rate the quality of this edited result? (From 1 to 10)
\begin{figure}[H]
  \centering
    % \vspace{-20pt}
    \includegraphics[width=0.8\linewidth]{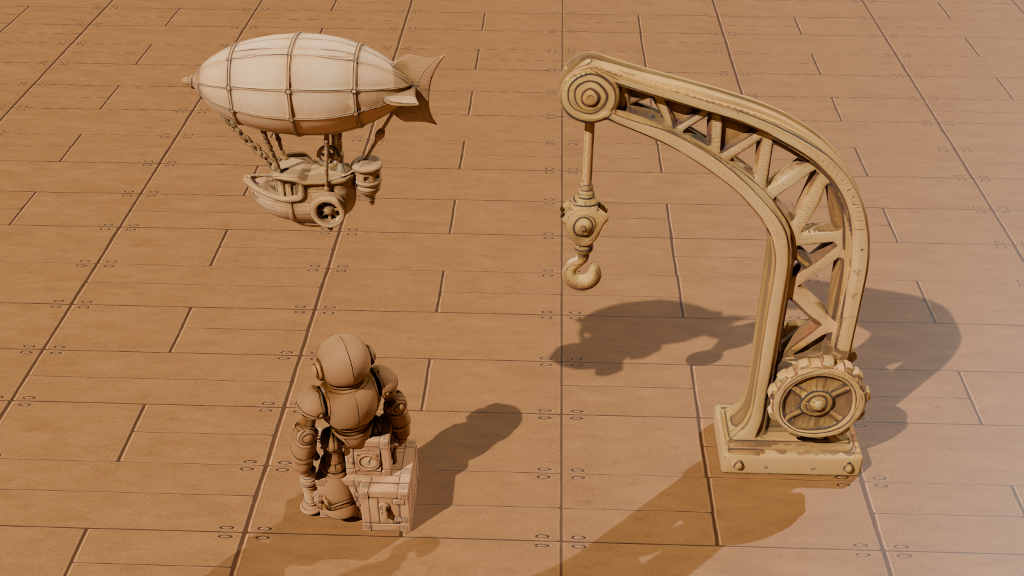}
    \caption*{Scene C}
    \vspace{-15pt}
\end{figure}
    \item How would you rate the quality of this edited result? (From 1 to 10)
\end{enumerate}

\subsection{Model Evaluation}
We adopt three VLM-based metrics to assess the spatial--semantic quality of generated scenes: \textit{Positional Coherency (Pos.)}, \textit{Rotational Coherency (Rot.)}, and \textit{Physically-grounded Semantic Alignment (PSA)}. 
These metrics are produced by GPT-4o following standardized rubrics. 
Below we provide the detailed definitions of each metric, complementing the high-level overview presented in Section 4.1.

\vspace{0.3em}
\textbf{Positional Coherency (Pos.).}
Positional Coherency measures how well the relative \emph{positions} of objects adhere to the spatial relations implied by the input description. 
The evaluator considers whether spatial relations such as left/right, front/back, and near/far are satisfied and whether violations disrupt the intended layout. 
A score in the range $[1, 100]$ is assigned based on a qualitative rubric assessing the overall positional consistency of the scene.

\vspace{0.3em}
\textbf{Rotational Coherency (Rot.).}
Rotational Coherency evaluates the coherence of object \emph{orientations} with directional cues in the textual input (e.g., facing or alignment constraints). 
The evaluator judges whether the facing directions are consistent with the intended semantics and whether misorientations meaningfully impact functional relationships. 
A score in the range $[1, 100]$ summarizes the global orientation coherence.

\vspace{0.3em}
\textbf{Physically-grounded Semantic Alignment (PSA).}
\textit{PSA} provides an overall assessment of the semantic correctness of the generated layout while explicitly incorporating physical feasibility. 
To compute \textit{PSA}, we use two components:
(i) \textit{Semantic Alignment (SA)}, a holistic $[1, 100]$ score capturing how well the overall spatial organization matches the intended description; and  
(ii) the \textit{Collision-Free ratio (CF)}, defined as the fraction of scenes without any inter-object collisions.
The final metric is computed as:
\begin{equation}
    \text{PSA} = \text{CF} \times \text{SA}.
\end{equation}
This formulation ensures that semantically correct but physically implausible layouts are appropriately penalized.

\vspace{0.3em}
For completeness, we provide the exact prompts (modified based on the prompts used by LayoutVLM) to obtain \textit{Pos.}, \textit{Rot.}, and \textit{SA}.

% We evaluate the spatial-semantic consistency of generated scenes using three sub-metrics: \textit{Positional Coherency (Pos.)}, \textit{Rotational Coherency (Rot.)}, and \textit{Physically-grounded Semantic Alignment (PSA)}. Our evaluation protocol is adapted from the prompting strategy introduced in LayoutVLM, with minor modifications to accommodate 3D scene layouts. Below, we include the exact prompt used to query GPT-4o for producing these evaluation scores.

% GPT-4o directly returns the scores for \textit{Pos.}, \textit{Rot.}, and \textit{SA}. To incorporate physical feasibility, we further adjust SA using the \textit{Collision-Free ratio (CF)} derived from our physical plausibility analysis. 

% The final physics-aware semantic metric, \textit{Physically-grounded Semantic Alignment (PSA)}, is computed as:
% \begin{equation}
%     \text{PSA} = \text{CF} \times \text{SA}.
% \end{equation}

% This formulation ensures that scenes achieving high semantic correctness but violating physical constraints (e.g., exhibiting object collisions) are penalized appropriately. The combination of semantic and physical reasoning provides a more reliable assessment of the overall spatial--semantic quality of generated scenes.

\noindent\rule{\linewidth}{1pt}
\vspace{-2em}
\captionof{listing}{Prompt - Positional Coherency (Pos.)}
\vspace{-0.7em}
\noindent\rule{\linewidth}{0.3pt}
\begin{lstlisting}[language=Python]
POSITION_PROMPT = """
You are a scene designer.
Given generated renderings of a 3D scene, your job is to evaluate how well an automated 3D layout generator performs.

The instruction given to the 3D layout generator was:

Evaluate the 3D layout generator as follows:
Assess the relative position (do not consider orientation) between assets: determine if related objects are placed near each other in a way that makes sense for their use.

Scoring Criteria for Position:
100-81: Excellently Positioned - Related objects are positioned near each other perfectly, facilitating their combined use.
80-61: Well Positioned - Most related objects are logically placed near each other, with few exceptions.
60-41: Adequately Positioned - Some related objects are not optimally placed, impacting their use together.
40-21: Poorly Positioned - Many related objects are placed far apart, hindering their joint use.
20-1: Very Poorly Positioned - Related objects are placed without consideration for their relationship, severely affecting functionality.
Please assess the image based on how coherently its layout aligns with the given target criteria.
Please provide justification and explanation for the score you give, in detail.
Always end your answer with:
### my final rating is: [replace this with a number between 1-100]
This is extremely important.
"""
\end{lstlisting}
\vspace{-0.7em}
\noindent\rule{\linewidth}{0.3pt}

\medbreak
\vspace{2em}
\noindent\rule{\linewidth}{1pt}
\vspace{-2em}
\captionof{listing}{Prompt - Rotational Coherency (Rot.)} 
\vspace{-0.7em}
\noindent\rule{\linewidth}{0.3pt}
\begin{lstlisting}[language=Python]
ORIENTATION_PROMPT = """
You are a scene designer.
Given generated renderings of a 3D scene, your job is to evaluate how well an automated 3D layout generator performs.

The instruction given to the 3D layout generator was:

Evaluate the 3D layout generator as follows:
Assess the coherency of asset orientation: Evaluate if related objects are oriented relative to each other in a way that makes sense for their use.

Scoring Criteria for Orientation:
100-81: Excellently Oriented - The orientation of objects perfectly complements their use and relationship with each other.
80-61: Properly Oriented - Most objects are oriented sensibly relative to each other, with minor misalignments.
60-41: Adequately Oriented - Several objects have orientations that do not fully support their use or relation.
40-21: Poorly Oriented - Many objects are oriented in ways that detract from their functionality or relation.
20-1: Very Poorly Oriented - Objects are oriented without any apparent logic, severely undermining their intended use and relationship.
Please assess the image based on how coherently its layout aligns with the given target criteria.
Please provide justification and explanation for the score you give, in detail.
Always end your answer with:
### my final rating is: [replace this with a number between 1-100]
This is extremely important.
"""
\end{lstlisting}
\vspace{-0.7em}
\noindent\rule{\linewidth}{0.3pt}

\noindent\rule{\linewidth}{1pt}
\vspace{-2em}
\captionof{listing}{Prompt - Semantic Alignment (PSA)}
\vspace{-0.7em}
\noindent\rule{\linewidth}{0.3pt}
\begin{lstlisting}[language=Python]
SA_PROMPT = """
You are a scene designer.
Given generated renderings of a 3D scene, your job is to evaluate how well an automated 3D layout generator performs.

The instruction given to the 3D layout generator was:

Evaluate the 3D layout generator as follows:

Layout criteria match: On a scale of 1 to 100, how well does the layout (i.e., just the layout, not the assets) capture the essence of the specified layout criteria?

Scoring Criteria:
100: Excellent - The layout of the scene (i.e., relative position and pose of objects in the scene) aligns very well with the criteria.
80: Good - The layout of the scene mostly aligns with the criteria (only ~10% of assets do not).
60: Ok - The layout of the scene somewhat aligns with the criteria (only ~30% of assets do not).
40: Poor - The layout of the scene (over ~50% of the assets) does not align with the criteria.
20: Very Poor - The layout of the whole scene does not capture the target criteria at all.

Please assess the image based on how coherently its layout aligns with the given target criteria.
Please provide justification and explanation for the score you give, in detail.
Always end your answer with:
### my final rating is: [replace this with a number between 1-100]
This is extremely important.
"""
\end{lstlisting}
\vspace{-0.7em}
\noindent\rule{\linewidth}{0.3pt}
\setcounter{figure}{0}

\section{Qualitative Examples}
In Figure \ref{fig9} and Figure \ref{fig10}, we present 18 additional scenes generated by \holodeck 2.0, each accompanied by the fine-grained input text and the rendered scene. In Figure \ref{fig11} and Figure \ref{fig12}, we also present 10 additional examples of scene editing, showing the original scene, the input editing instruction, and the final edited scene.

\begin{figure*}[!t]
  \centering
    \includegraphics[width=\linewidth]{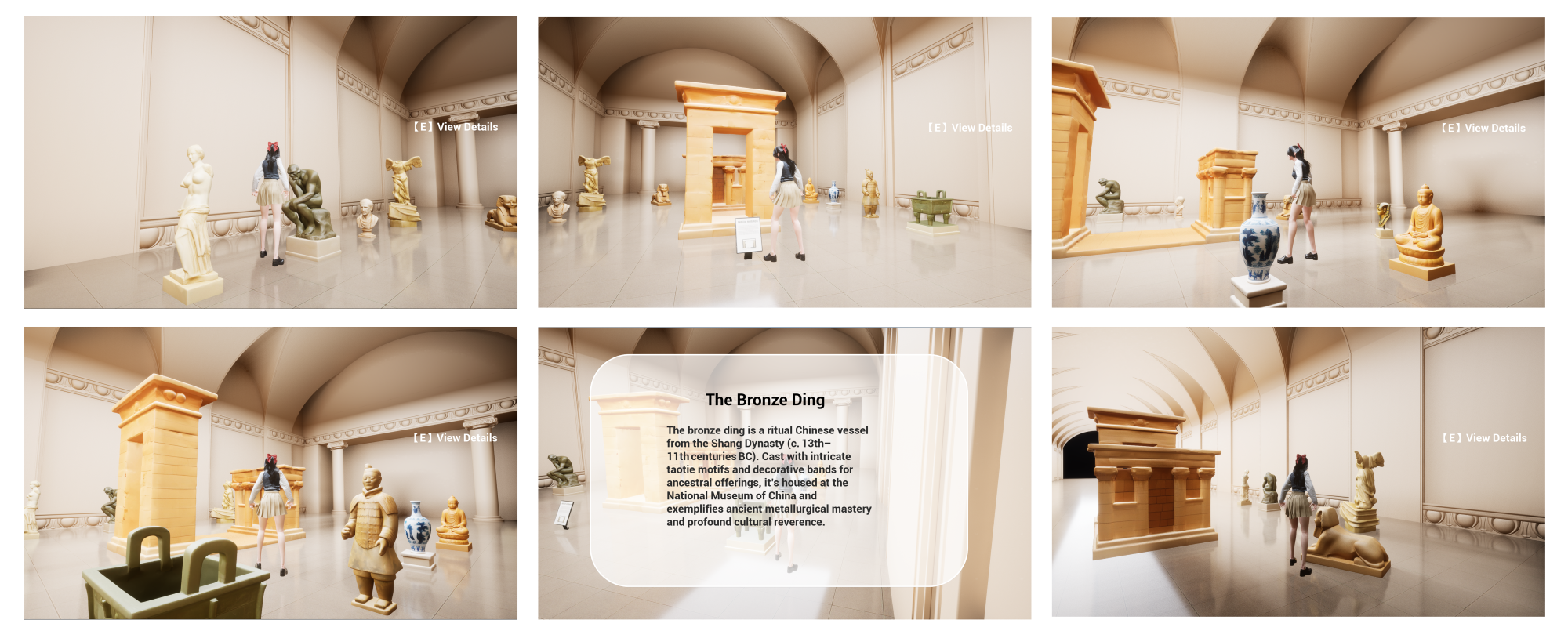}
    \caption{In-game interface of the 3D scene generated by \holodeck 2.0}
    \label{fig8}
\end{figure*}

\begin{figure*}[!t]
  \centering
    \includegraphics[width=\linewidth]{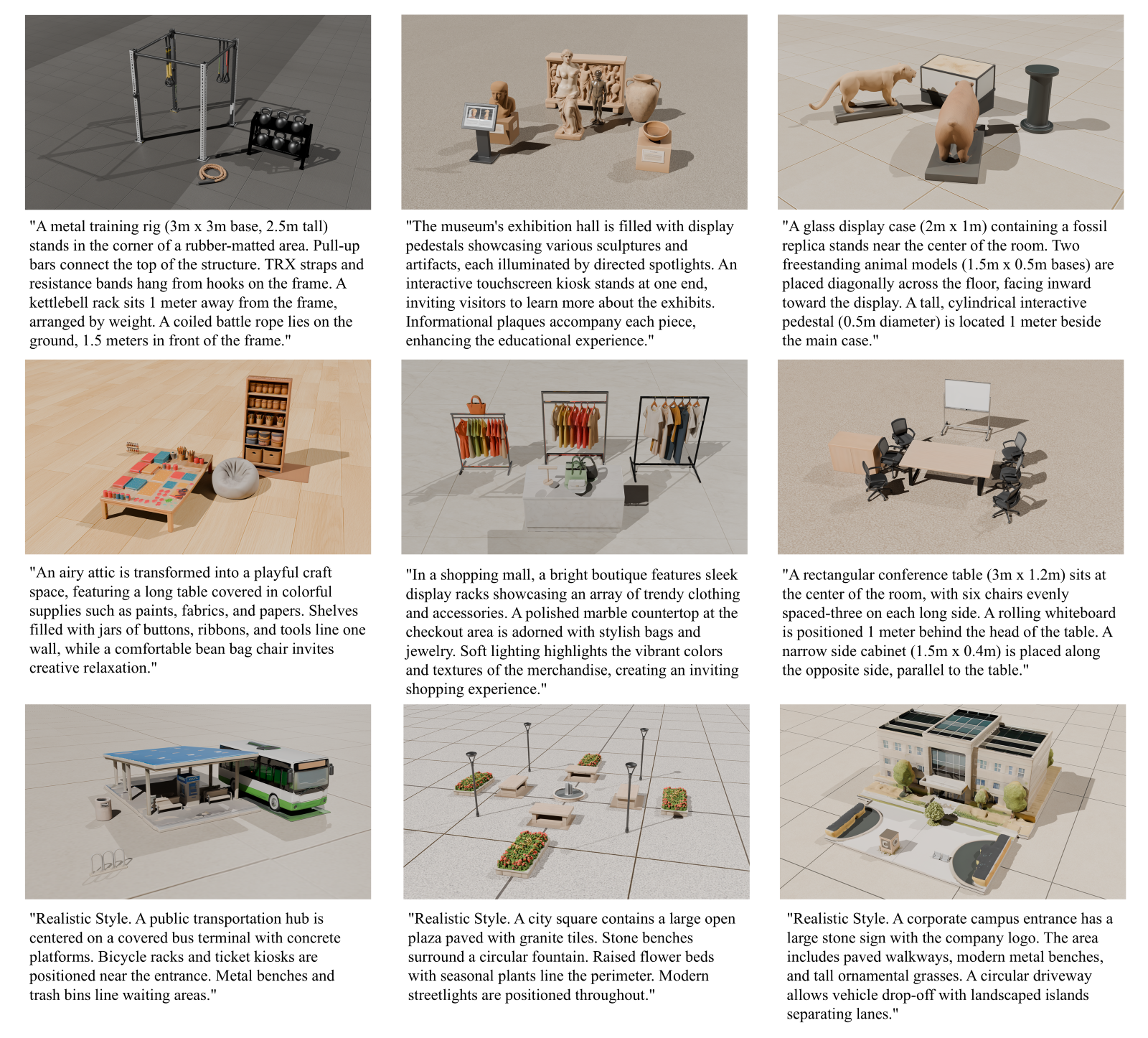}
    \caption{Scenes generated by \holodeck 2.0 and the corresponding fine-grained input text}
    \label{fig9}
\end{figure*}
\begin{figure*}[!t]
  \centering
    \includegraphics[width=\linewidth]{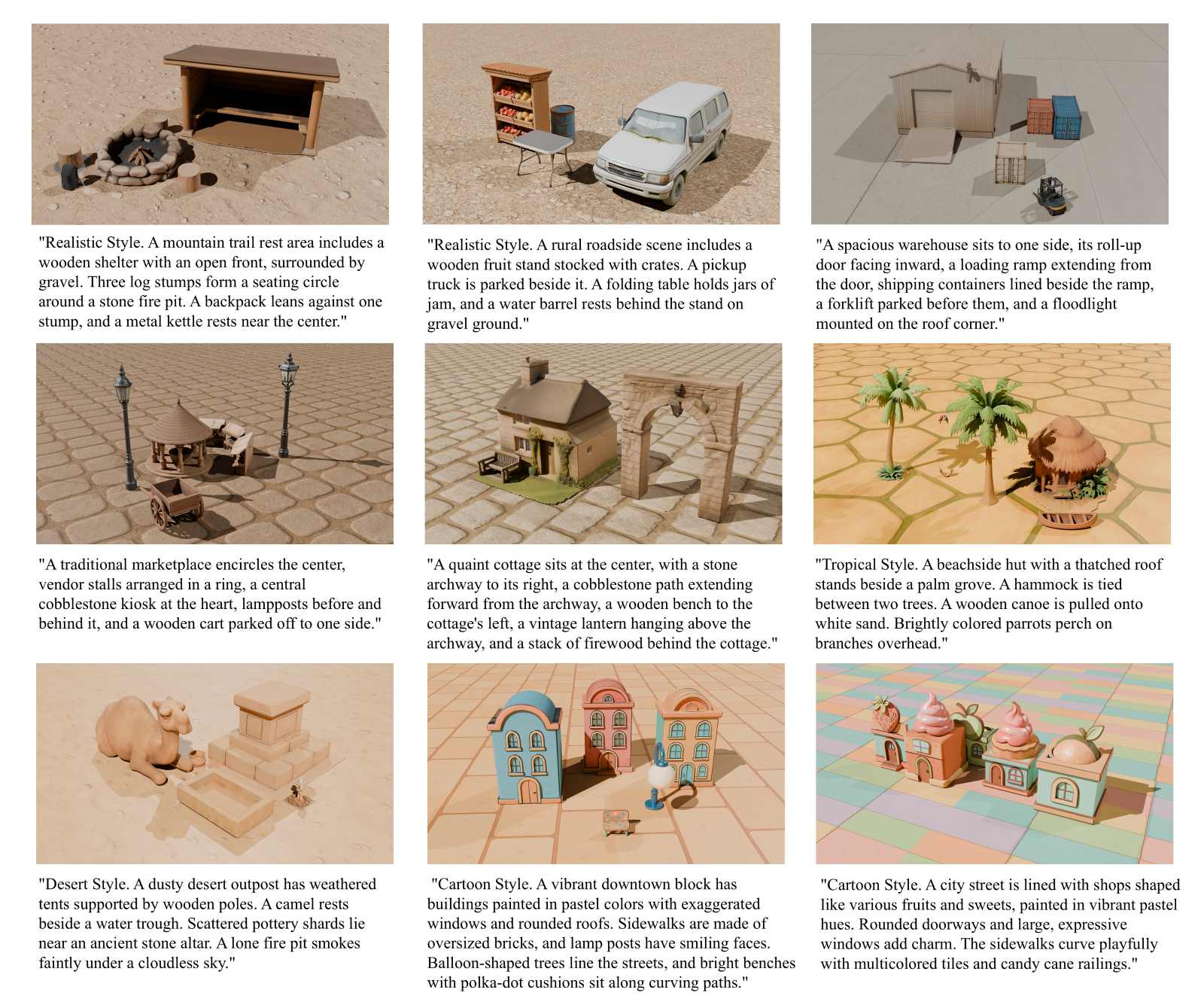}
    \caption{Scenes generated by \holodeck 2.0 and the corresponding fine-grained input text}
    \label{fig10}
\end{figure*}
\begin{figure*}[!t]
  \centering
    \includegraphics[width=\linewidth]{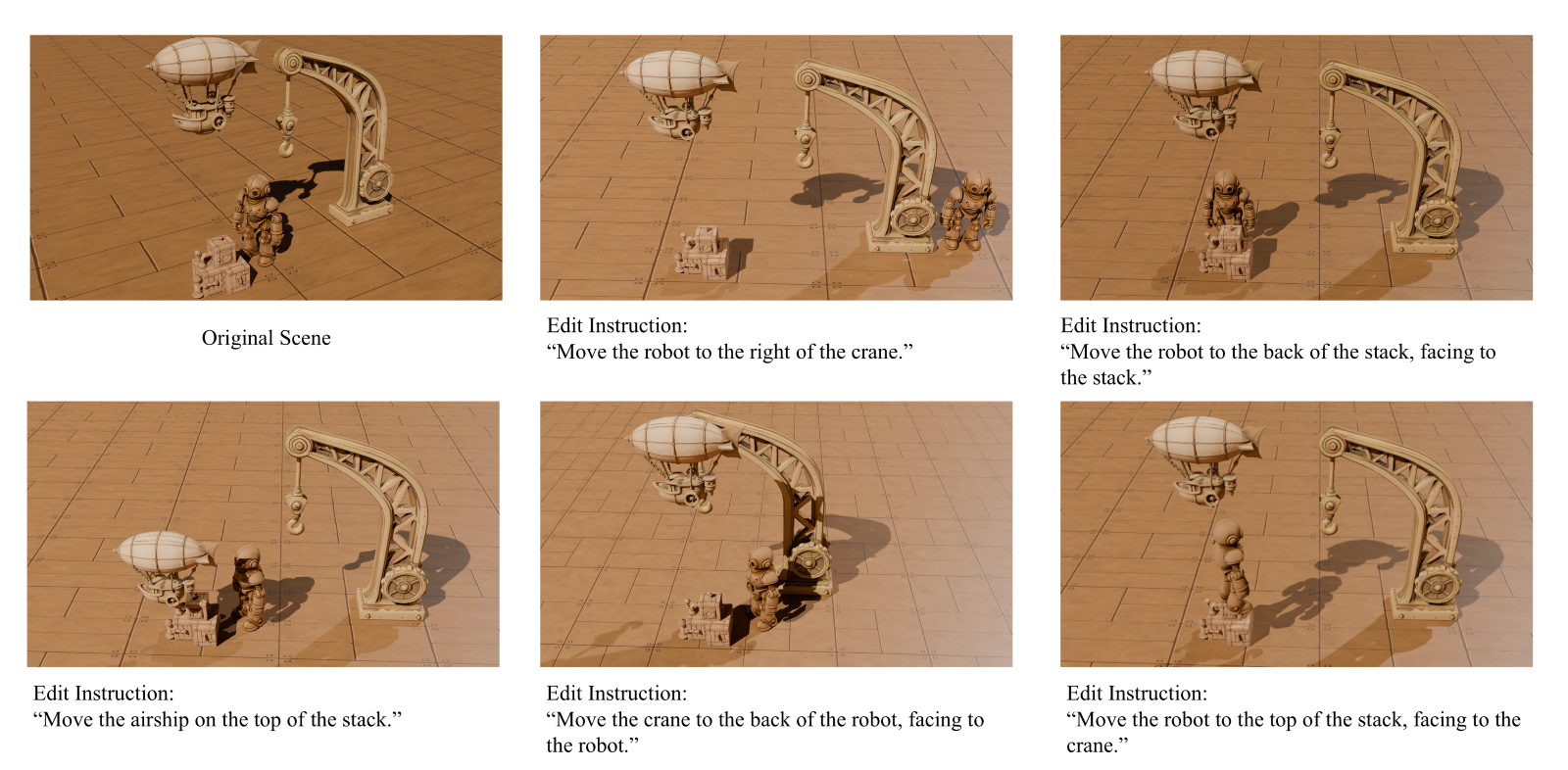}
    \caption{Edited scenes produced by the Scene Editing Module using the instructions}
    \label{fig11}
\end{figure*}
\begin{figure*}[!t]
  \centering
    \includegraphics[width=\linewidth]{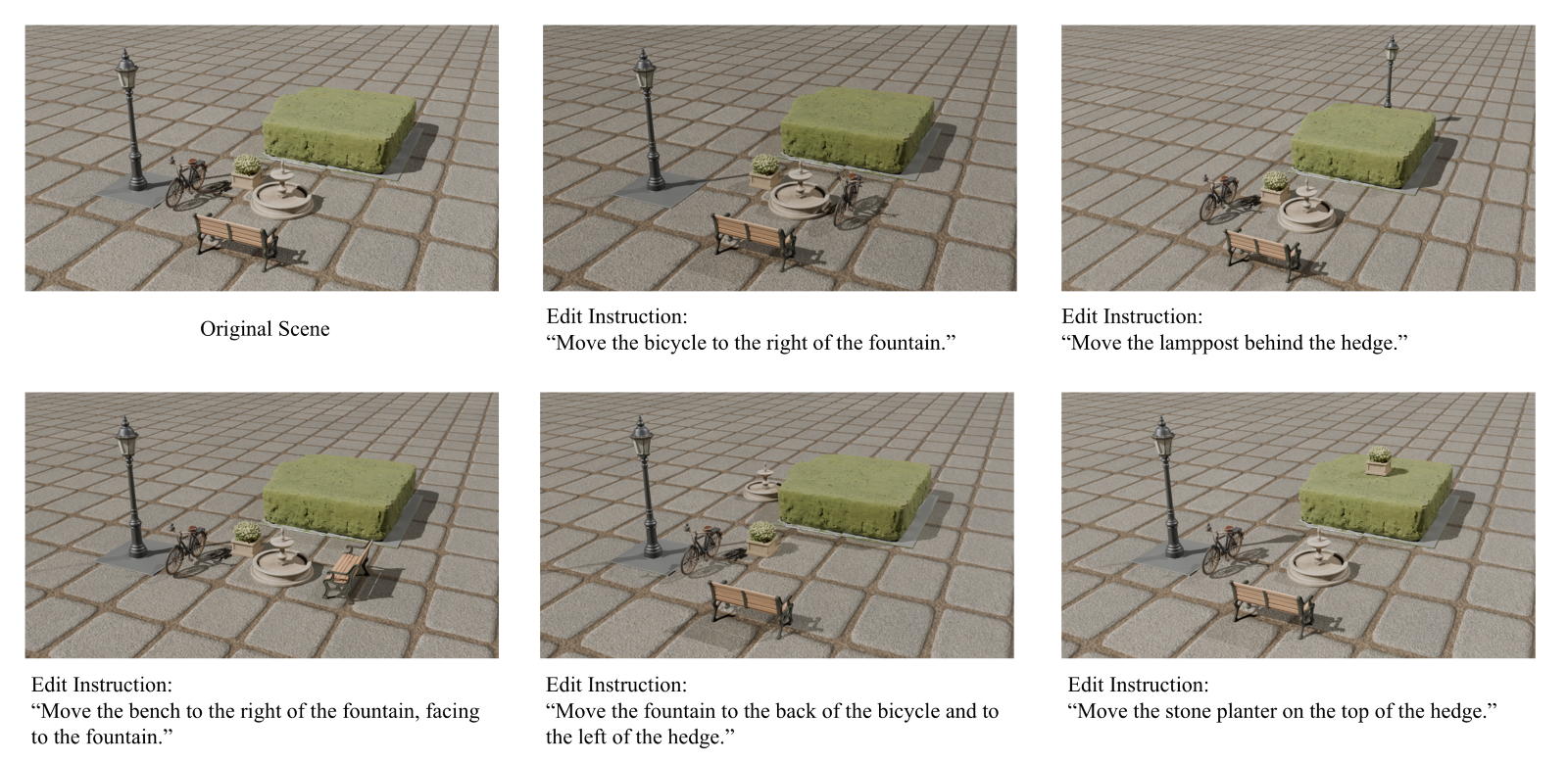}
    \caption{Edited scenes produced by the Scene Editing Module using the instructions}
    \label{fig12}
\end{figure*}

\end{document}